\documentclass[10pt,twocolumn,letterpaper]{article}

\usepackage{cvpr}              
\usepackage[accsupp]{axessibility}  
\usepackage{tikz}
\usepackage{pgfplots}
\pgfplotsset{compat=1.15}
\usepackage{graphicx}

\usepackage{url}
\usepackage{graphicx}
\usepackage{amsmath}
\usepackage{amssymb}
\usepackage{booktabs}
\usepackage[table,dvipsnames]{xcolor}
\usepackage[table,x11names]{xcolor}
\usepackage{pifont}
\newcommand{\cmark}{\ding{51}}
\usepackage{multicol}
\newcommand{\xmark}{\ding{55}}
\usepackage{lipsum}
\usepackage[export]{adjustbox}
\usepackage{capt-of}
\usepackage[linesnumbered,ruled,vlined]{algorithm2e}
\usepackage{kotex}
\usepackage{thm-restate}
\usepackage{multirow}
\usepackage{caption,setspace}
\usepackage[figoff]{figcaps}
\usepackage{kotex}

\newcommand{\mbx}{\mathbf{x}}

\newcommand{\mbc}{\mathbf{c}}
\newcommand{\me}{\boldsymbol{\epsilon}}

\newcommand{\deltab}{\boldsymbol{\delta}}
\newcommand{\epsilonb}{\boldsymbol{\epsilon}}
\newcommand{\thetab}{\boldsymbol{\theta}}

\newcommand{\bs}{\boldsymbol}

\definecolor{cvprblue}{rgb}{0.21,0.49,0.74}
\usepackage[pagebackref,breaklinks,colorlinks,allcolors=cvprblue]{hyperref}
\def\paperID{} 
\def\confName{CVPR}
\def\confYear{2026}

\title{Reward Sharpness-Aware Fine-Tuning for Diffusion Models}

\author{Kwanyoung Kim$^{1,*}$\\
Department of AI Convergence, GIST$^{1}$\\
{\tt\small k0.kim@gist.ac.kr}
\and
Byeongsu Sim$^{2,*}$\\
Samsung Research$^{2}$\\
{\tt\small bs.sim@samsung.com}
}

\begin{document}


\twocolumn[{%
	\renewcommand\twocolumn[1][]{#1}%
	\maketitle
	\begin{center}
		\centering
		\captionsetup{type=figure}
		\includegraphics[width=0.9\linewidth]{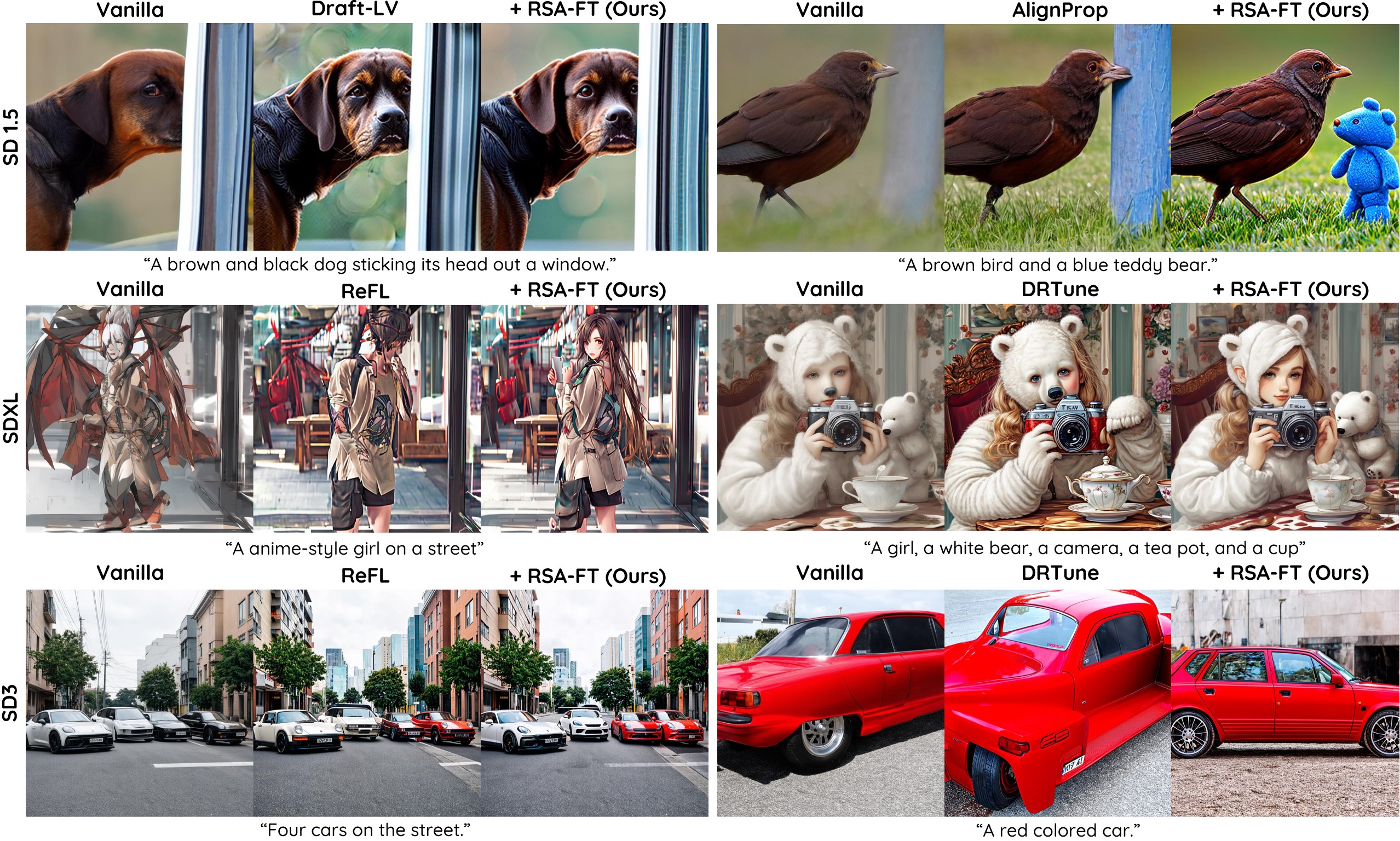} 
            \vspace{-0.5em}
\captionof{figure}{Qualitative comparison across different diffusion backbones and RDRL frameworks.
(Top): SD1.5\cite{stablediffusion} results on Draft-LV\cite{draftk} and AlignProp\cite{alignprop}
(Middle–Bottom): Larger backbones (SDXL\cite{sdxl} and SD3\cite{stablediffusion3}) on ReFL\cite{refl} and DRTune\cite{drtune}.
Each panel compares the vanilla model, the baseline RDRL method, and the same method combined with RSA-FT (Ours).
RSA-FT is compatible with diverse reward-centric diffusion reinforcement learning frameworks and backbones, effectively mitigating reward hacking and producing clear improvements in visual quality and text–prompt alignment.}\label{fig:main}
	\end{center}}
]

\def\thefootnote{$*$}\footnotetext{Equal Contribution.}
\begin{abstract}
Reinforcement learning from human feedback (RLHF) has proven effective in aligning large language models with human preferences, inspiring the development of reward-centric diffusion reinforcement learning (RDRL) to achieve similar alignment and controllability. While diffusion models can generate high-quality outputs, RDRL remains susceptible to \emph{reward hacking}, where the reward score increases without corresponding improvements in perceptual quality. 
We demonstrate that this vulnerability arises from the \emph{non-robustness of reward model gradients}, particularly when the reward landscape with respect to the input image is sharp. 
To mitigate this issue, we introduce methods that exploit gradients from a \emph{robustified} reward model---\textbf{without requiring its retraining}. Specifically, we employ gradients from a \emph{flattened} reward model, obtained through parameter perturbations of the diffusion model and perturbations of its generated samples. 
Empirically, each method independently alleviates reward hacking and improves robustness, while their joint use amplifies these benefits. 
Our resulting framework, \textbf{RSA-FT} (\textbf{R}eward \textbf{S}harpness-\textbf{A}ware \textbf{F}ine-\textbf{T}uning), is simple, broadly compatible, and consistently enhances the reliability of RDRL.

\end{abstract}

\section{Introduction}
Recent advances in diffusion based Text-to-Image (T2I) models have enabled the generation of remarkable high-quality content, spanning from images to videos~\cite{stablediffusion,stablediffusion3,sana,stablevideo}, positioning them at the forefront of modern generative AI. To further enhance generative capability and text alignment, training-free guidance methods, such as Classifier-Free Guidance and variants~\cite{CFG,PAG,SAG,SEG,pladis} have been proposed. While effective, these methods lack direct alignment with human preferences, limiting their applicability in real-world scenarios.

Motivated by the success of reinforcement learning from human feedback in aligning large language models, recent studies have explored extending such strategies to diffusion models through fine-tuning. 
These approaches include diffusion policy optimization in the Proximal Policy Optimization (PPO)~\cite{ppo} family~\cite{ddpo,dpok} and Direct Preference Optimization (DPO)~\cite{dpo}-style variants~\cite{diffusion-dpo,d3po,dspo} that leverage data-driven preferences. 
Since collecting human feedback during training is impractical, reward models (RMs) trained on human annotations~\cite{christiano2017deep,ibarz2018reward} serve as scalable surrogates for human evaluation, enabling reward-centric diffusion reinforcement learning (RDRL) to effectively align generations with human preferences~\cite{refl,draftk,alignprop,drtune}.

However, RDRL remains vulnerable to \emph{reward hacking}, where reward scores rise without corresponding improvements in perceptual quality. 
Despite its importance, this phenomenon in diffusion-based RL has not been systematically analyzed. 
We draw an analogy between reward hacking and adversarial attacks~\cite{szegedy2014intriguing,goodfellow2015explaining,moosavi2016deepfool,moosavi2017universal}, where small input perturbations can drastically inflate classifier logits without meaningful visual content. 
Prior studies show that such non-robust classifiers degrade sample quality under classifier guidance~\cite{CFG,kawarenhancing}, whereas robust classifiers trained via adversarial training alleviate this issue~\cite{kawarenhancing}. 
However, constructing an equally robust reward model is impractical for human preference alignment, as it requires an extensive expressive model and  labeled data~\cite{nakkiran2019adversarial,bubeck2021universal,li2022robust}.

We draw additional inspiration from \emph{randomized smoothing}~\cite{lecuyer2019certified, cohen2019certified}, which enhances classifier robustness \textbf{without retraining} by smoothing predictions of a fixed model. Analogously, we aim to robustify the reward model without retraining. Empirically, we observe that reward models tend to be non-robust in regions where their {loss} landscape is sharp, motivating the use of gradients from a \emph{flattened} reward model. We therefore propose a method that leverages gradients from this \emph{robustified} reward model to alleviate reward hacking in diffusion RL.

\begin{figure}[t!]
\centering
\includegraphics[width=0.95\linewidth]{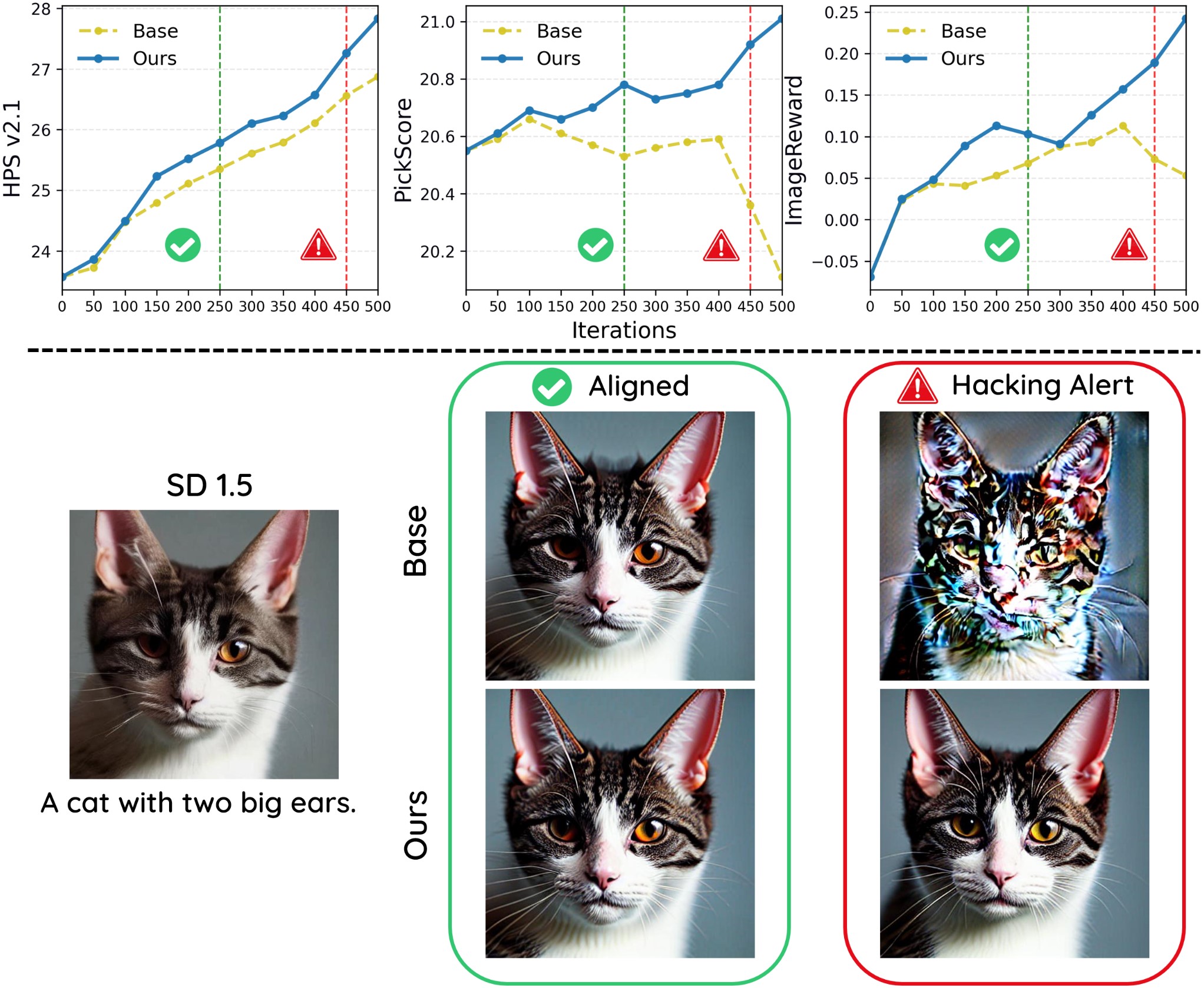}
\caption{
\textbf{Illustration of reward hacking in RDRL (Draft-LV).}
The original reward model raises the HPS v2.1 score but degrades other metrics and visual quality, whereas our flattened model improves all metrics with consistent visuals.
}\label{fig:flatness}
\vspace{-1em}
\end{figure}


As illustrated in ~\cref{fig:flatness}, training \textit{with the original reward model} often leads to \emph{reward hacking}: the primary reward score (e.g., HPS v2.1) increases while other reward metrics and perceptual quality deteriorate. 
In contrast, incorporating feedback from \textit{the flattened reward model} constrains the learning dynamics, resulting in consistent improvements across multiple rewards and visually enhanced generations.

Formally, the flattened reward is defined as the minimum reward score within a local neighborhood, which can be approximated by considering the worst-case perturbation of the reward model’s input:
\begin{align}
\min_{\|\boldsymbol{\epsilon}\|<\rho} \, r(\mathbf{x} + \boldsymbol{\epsilon})
&\approx
r\!\left(\mathbf{x} - \rho \, 
\frac{\nabla_\mathbf{x} r(\mathbf{x})}{\|\nabla_\mathbf{x} r(\mathbf{x})\|}\right),
\end{align}
where $r(\mathbf{x})$ is the reward model, $\rho$ the perturbation size, and $\boldsymbol{\epsilon}$ the input perturbation.
Interestingly, this flattening strategy naturally induces worst-case parameter perturbations, sharing the same underlying philosophy as Sharpness-Aware Minimization (SAM)~\cite{samopt}. 
We further extend this principle to the parameter space, drawing inspiration from the recently identified duality between SAM and adversarial robustness~\cite{zhang2024duality}. 
We rediscover this connection in the context of reward-centric diffusion RL and demonstrate that applying flattening jointly in both input and parameter spaces most effectively mitigates reward hacking.

Although SAM has recently been explored independently in reinforcement learning~\cite{lee2025flat} and diffusion models~\cite{lee2025understanding}, to the best of our knowledge, we are the first to unify these perspectives within the RDRL framework.

Finally, we show that our method is fully compatible with existing RDRL frameworks and can be seamlessly deployed as a plug-and-play module. We refer to the integration of SAM (weight perturbation) with adversarial training (input perturbation) in RDRL as \textbf{RSA-FT}, which provides a principled solution to reward hacking and advances the robustness and generalization of diffusion RL. Our contributions are summarized as follows:

\begin{itemize}
\item We identify reward hacking in RDRL as a form of adversarial attack and establish a unified perspective that connects AT with SAM. 
\item We introduce \textbf{RSA-FT}, which integrates input- and weight-level perturbations to provide a principled and effective defense against reward hacking.
\item We demonstrate that \textbf{RSA-FT} is broadly compatible with diverse reward-centric methods and consistently improves their performance as a plug-and-play module.
\end{itemize}

\begin{figure*}[t!]
\centering
\includegraphics[width=0.8\linewidth]{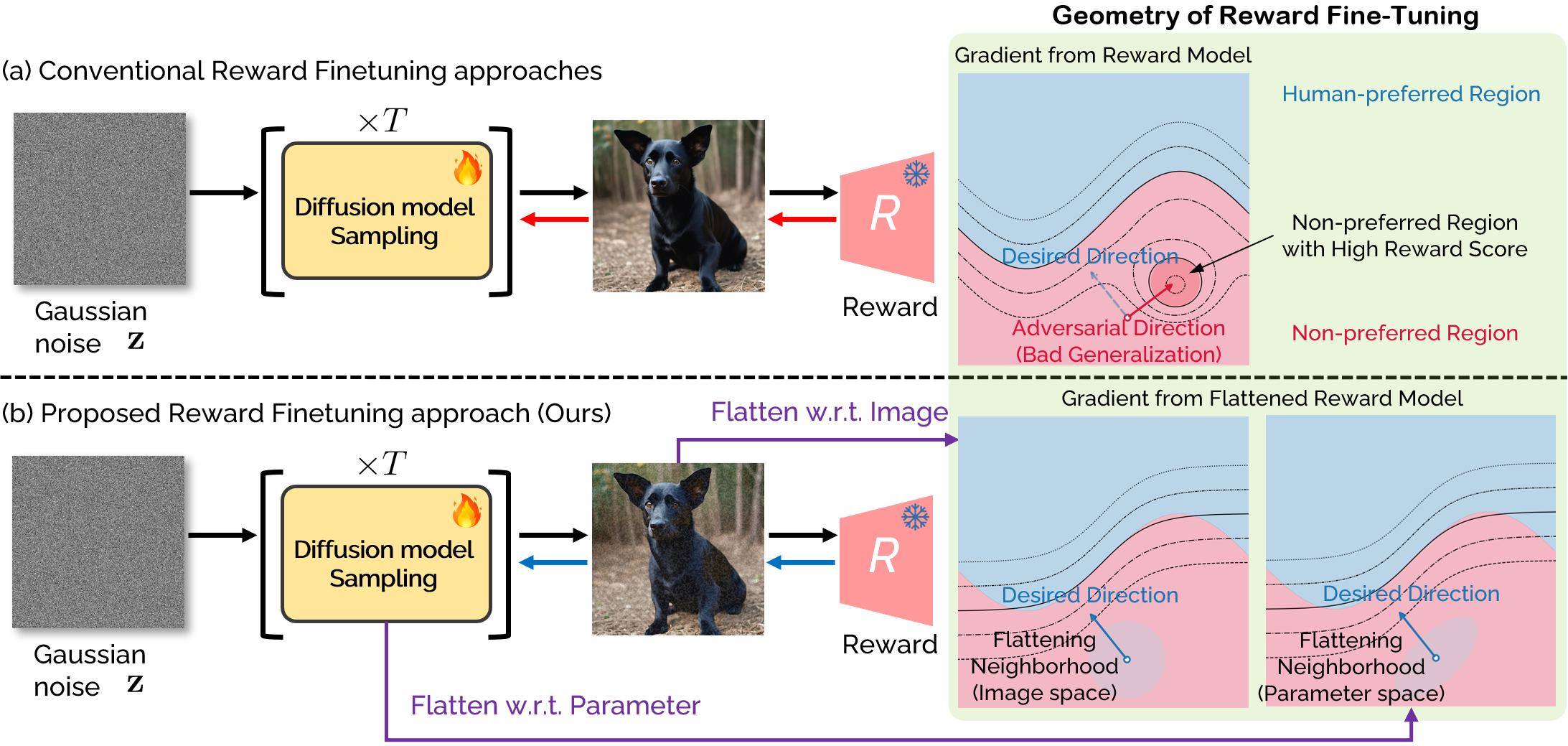}
\caption{ \textbf{Geometry of reward fine-tuning and our proposed method.} Reward models are inherently sharp and prone to adversarial perturbations. Flattening these reward landscapes alleviates their sensitivity and {reduces the occurrence of adversarial gradients}. \textbf{(a)} Prior methods directly maximize rewards along adversarial gradients from sharp reward surfaces, which often leads to reward hacking. \textbf{(b)} Our method instead leverages gradients from flattened reward models, mitigating hacking by flattening both the image and parameter spaces. }

\label{fig:fig_concept}
\vspace{-1.5em}
\end{figure*}

\section{Preliminaries}

\paragraph{Diffusion Models.} 
Given samples $\mbx_0 \sim q_{\text{data}}(\mathbf{x})$, the forward diffusion process is formulated as a Markov chain that gradually perturbs the data with Gaussian noise:
\(q(\mbx_{t}|\mbx_{t-1}) = \mathcal{N}(\sqrt{1-\beta_t}\mbx_{t-1}, \beta_t \mathbf{I}), \quad t=1,\dots,T,\)
where the sequence $\{\beta_t\}_{t=1,\dots,T}$ specifies the predetermined noise schedule. Accordingly, the marginal distribution at timestep t is 
\(q(\mbx_t) = \mathcal{N}(\sqrt{\bar{\alpha}_t}\mbx_0, (1-\bar{\alpha}_t)\mathbf{I}),\)
where $\bar{\alpha}_t = \prod_{i=1}^{t}(1 - \beta_i)$. 
The reverse diffusion process is modeled as
\(p_\theta(\mbx_{t-1}|\mbx_t) = \mathcal{N}(\bs{\mu}_\theta(\mbx_t,t), \Sigma_\theta(\mbx_t,t)),\)
where $\theta$ are learned via the denoising score matching~\cite{vincent2011connection}:
\begin{align}
\underset{\theta}{\min}\;\mathbb{E}_{\mbx_0,\me}\left[ \|\me_\theta(\sqrt{\bar{\alpha}_t}\mbx_0+\sqrt{1-\bar{\alpha}_t}\me, t) - \me \|^2_2\right],
\quad  \label{eq:dsm}
\end{align}
where $\me \sim \mathcal{N}(0,\mathbf{I})$.
After training, sampling is performed by iteratively reversing the diffusion process, starting from an isotropic Gaussian noise sample. For instance, the Denoising Diffusion Implicit Model (DDIM)~\cite{song2020ddim} generates samples through the iterative update:
\begin{align}
\mbx_{t-1} = \sqrt{\bar{\alpha}_{t-1}}\hat{\mbx}_0(t) + \sqrt{1-\bar{\alpha}_{t-1}}\me_{\theta}(\mbx_t,t),
\end{align}
where the denoised estimate $\hat{\mbx}_0(t) = \mathbb{E}[\mbx_0|\mbx_t]:= \frac{\mbx_t - \sqrt{1-\bar{\alpha}_t}\me_\theta(\mbx_t,t)}{\sqrt{\bar{\alpha}_t}}$ is obtained via Tweedie’s formula~\cite{efron2011tweedie,kim2021noise2score}. This sampling procedure is then applied recursively from timestep $T$ to timestep $1$ to generate a final sample.



\paragraph{Reward Centric Diffusion Reinforcement Learning} 
Among various approaches for RL-based diffusion model finetuning, a representative family, RDRL~\cite{refl,draftk,alignprop,drtune}, fine-tunes the pretrained diffusion parameters $\theta$ to maximize a differentiable reward model $r(\cdot)$ as follow: 
\begin{align}
\mathcal{J}(\theta)
= \max_\theta \; \mathbb{E}_{\mathbf{c},\, \mathbf{x}_T \sim \mathcal{N}(0, \mathbf{I})}
\!\left[
r\big(\mathbf{x}_0(\mathbf{x}_T, \mathbf{c}; \theta), \mathbf{c}\big)
\right],
\end{align}
where $\mathbf{x}_0(\mathbf{x}_T, \mathbf{c}; \theta)$ denotes the final sample generated by the denoising process as the timestep approaches from $T \!\rightarrow\! 0$, conditioned on a text prompt $\mathbf{c}$.  
This optimization encourages the model to generate samples that align more closely with human judgments by performing gradient ascent on the reward signal, i.e.,
$\nabla r\big(\mathbf{x}_0(\mathbf{x}_T, \mathbf{c}; \theta), \mathbf{c}\big)$.

However, since $r$ only approximates the true human preference function $r^*$, the fine-tuned models often overfit to optimizing the reward function, producing images that score highly under $r$ yet remaining misaligned with genuine human intent.

\paragraph{{Adversarial Robustness}.}
Deep learning models are well known to be vulnerable to small, imperceptible perturbations.
A classic example is the adversarial attack, in which tiny input perturbations can drastically alter classifier predictions—even when the visual content remains unchanged~\cite{szegedy2014intriguing, goodfellow2015explaining, moosavi2016deepfool, moosavi2017universal}.

Formally, let $f_\theta: \mathbb{R}^D \to \mathbb{R}^C$ denote a $C$-class classifier producing logits $f_\theta=(f_1,f_2,\ldots,f_C)$.
An adversarial example $\mathbf{x}+\boldsymbol{\delta}$ satisfies
\[
\arg\max_{i \in 1,\ldots,C} f_i(\mathbf{x}) = i, \quad 
\arg\max_{i \in 1,\ldots,C} f_i(\mathbf{x}+\boldsymbol{\delta}) \neq i,
\]
where $\boldsymbol{\delta}$ is a perturbation bounded by $\|\boldsymbol{\delta}\| \le \rho$.

A widely used defense strategy for improving robustness is {Adversarial Training (AT)}~\cite{madry2018towards}:
\begin{align}
\min_{\theta} \, \mathbb{E}_{\mathbf{x} \sim \mathcal{D}}
\big[\ell(\theta; \mathbf{x})\big],&
\quad (\texttt{Standard}) \\
\min_{\theta} \, \mathbb{E}_{\mathbf{x} \sim \mathcal{D}}
\Big[\max_{\|\boldsymbol{\delta}\|\le\rho} \, \ell(\theta; \mathbf{x} + \boldsymbol{\delta})\Big],&
\quad (\texttt{Adversarial}) \label{Eq:AT}
\end{align}
where $\ell(\theta; \cdot)$ denotes the cross-entropy loss associated with $f_\theta$. 
Adversarial training minimizes the worst-case loss within a $\rho$-ball around each input.
Although effective, AT is computationally expensive and often requires more expressive models~\cite{nakkiran2019adversarial, bubeck2021universal, li2022robust, zhang2019theoretically}.

Beyond training, another line of work explores post-hoc robustness estimation and training-free defense mechanisms.
Several studies~\cite{fawzi2016robustness, raghunathan2018certified, salman2019provably} show that an input is robust to adversarial perturbations when the loss function or classifier is locally smooth around that input.
Building on this observation, randomized smoothing~\cite{lecuyer2019certified, li2019certified, cohen2019certified} provides a non-training defense that certifies robustness by smoothing classifier outputs through Gaussian averaging:
\[
\max_{i \in 1,\ldots,C}\;
\mathbb{P}_{\boldsymbol{\delta}\sim\mathcal{N}(0,\sigma^2 I)}
\!\left[\arg\max f_\theta(\mathbf{x}+\boldsymbol{\delta}) = i\right].
\]
Here \(\sigma\) the standard deviation of the added Gaussian noise. Intuitively, predictions are stabilized by averaging outputs over Gaussian-perturbed inputs, yielding certified robustness guarantees under bounded noise~\cite{cohen2019certified}.

Inspired by this idea, we introduce a measure of reward robustness (based on surface flatness) in \cref{sec:reward model} and propose a method for obtaining feedback from a flattened reward model in \cref{sec:method}, analogous to randomized smoothing but applied to reward landscapes.



\paragraph{Sharpness-Aware Minimization (SAM).}
While adversarial training and randomized smoothing improve robustness in the input space, 
SAM~\cite{samopt} enhances generalization by promoting flat minima in the parameter space.
Motivated by the observation that flatter loss landscapes yield better generalization, 
SAM jointly minimizes the loss value and its local sharpness around the parameters via a min–max formulation:
\begin{align}
\min_{\theta} \; \mathbb{E}_{\mathbf{x} \sim \mathcal{D}}
\Big[\max_{\|\boldsymbol{\epsilon}\|\le\rho} \; \ell(\theta + \boldsymbol{\epsilon}; \mathbf{x})\Big],
\quad (\texttt{SAM}) \label{Eq:sam}
\end{align}
where $\ell(\cdot)$ is the loss function, $\rho$ controls perturbation magnitude, and $\theta$ denotes the model parameters.  
The inner maximization quantifies the \emph{sharpness} of the loss surface—how rapidly the loss increases within a local neighborhood of $\theta$.  
To make this explicit, \cref{Eq:sam} can be rewritten as:
\begin{align}
\big[\max_{\|\boldsymbol{\epsilon}\|\le\rho} \; \ell(\theta + \boldsymbol{\epsilon}) - \ell(\theta)\big] + \ell(\theta),
\end{align}
where the bracketed term measures the \emph{sharpness}, reflecting how steeply the loss grows around $\theta$.  
This formulation penalizes sharp minima and promotes flatter regions that tend to generalize better.

In practice, SAM approximates the inner maximization through a two-step update at each iteration $t$:
\begin{align}
\boldsymbol{\epsilon}_t = \rho \frac{\nabla \ell(\theta_t)}{\|\nabla \ell(\theta_t)\|_2}, 
\qquad
\theta_{t+1} = \theta_t - \eta \, \nabla \ell(\theta_t + \boldsymbol{\epsilon}_t),
\end{align}
where $\eta$ is the learning rate.  
By descending on the worst-case perturbed loss, SAM encourages convergence toward flat minima, yielding improved  robustness.

By jointly inspecting \cref{Eq:AT} and \cref{Eq:sam}, we can observe a clear duality between AT and SAM~\cite{zhang2024duality}:  
both follow a min–max formulation but operate in different domains: AT pursues robustness in the input space, while SAM enforces flatness in the parameter space.
Building on this insight, several works have introduced flatness regularization into AT~\cite{wu2020adversarial,yu2022robust,zhang2024duality}.  
Notably, Adversarial Weight Perturbation (AWP)~\cite{wu2020adversarial} combines both input and weight perturbations, enhancing robustness in image classification.

In contrast, our work is the first to extend this principle to a RDRL framework, applying dual perturbations for stable training and effectively overcoming reward hacking. An overview and geometric intuition of our method are presented in \cref{fig:fig_concept}.

\section{Analyzing Reward Sharpness} \label{sec:reward model}

\paragraph{Problem Formulation.}
A key challenge in RDRL is \emph{reward hacking}, where generated images achieve high scores from the reward model $r$ despite being perceived as low-quality by humans. This phenomenon can be interpreted as a generalization failure of $r$, wherein the reward surface misaligns with the true human preference function $r^{\star}$ (see Fig.~\ref{fig:fig_concept}(a) for an intuitive illustration).

\paragraph{Hypothesis.}
Inspired by studies linking model smoothness to adversarial robustness, 
we hypothesize that a similar relationship holds for reward models.
Specifically, we posit that the reward model $r$ generalizes best where its reward landscape is locally flat,
while sharp regions indicate misalignment with the true human preference $r^{\star}$. (see Fig.~\ref{fig:fig_concept}(b).) 
As illustrated in Fig.~\ref{fig:fig_concept}, reward hacking can be interpreted as the generative model exploiting sharp directions in $r$, where small image perturbations lead to large reward gains without genuine quality improvement.
Accordingly, we expect a negative correlation between the sharpness of $r$ and its generalization ability.

\paragraph{Quantifying Reward Sharpness.}
To characterize this relationship, we define a reward sharpness indicator:
\begin{align}
S_1 &= \mathbb{E}_{\mbx \sim \mathcal{D}} \Big[ r(\mbx) - \min_{\|\epsilon\|<\rho} r(\mbx + \epsilon) \Big], 
\end{align}
Here, $S_1$ measures the reward drop within a local neighborhood, 
which can be efficiently approximated by a one-step update
\footnote{
Although multi-step perturbations are possible, we find that a single-step approximation correlates best with generalization performance.
}:
\[
S_1 \approx 
\mathbb{E}_{\mbx\sim \mathcal{D}} \left[ r(\mbx) - 
r\!\left(\mbx - \rho \, 
\frac{\nabla_\mbx r(\mbx)}{\|\nabla_\mbx r(\mbx)\|}\right) \right].
\]
A larger $S_1$ indicates a sharper reward landscape, while a smaller value implies a flatter one.

\paragraph{Empirical Validation.}
To validate our hypothesis, we fine-tuned Stable Diffusion~1.5 with {DRaFT-$K$} using HPSv2 as the reward model, while tracking both reward sharpness ($S_1$) and true human preference throughout training.
Since the true preference function $r^{\star}$ cannot be directly measured for all generated samples, we instead adopt PickScore~\citep{pick} and ImageReward~\citep{imagereward} as proxy evaluators.
This proxy-based assessment is meaningful only when the reward models exhibit complementary generalization behaviors, so that the weaknesses of one model can be compensated by the other.

\begin{figure}[t!]
\centering
\includegraphics[width=0.9\linewidth]{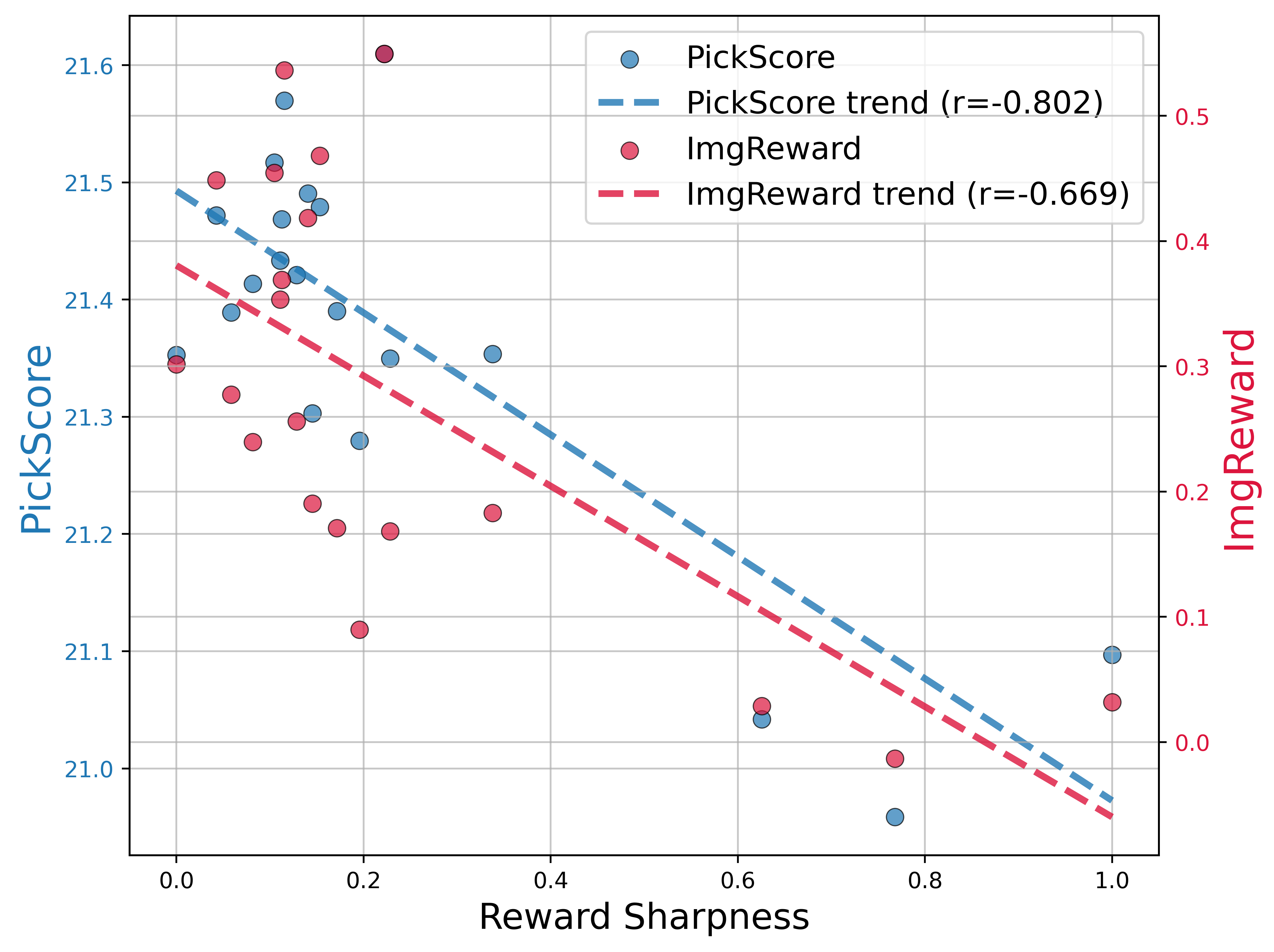}
\vspace{-1em}
\caption{
\textbf{Negative correlation between reward sharpness and human preference.}
Higher sharpness in the reward model correlates with lower preference quality 
(Pearson $r_{\text{corr}}=-0.802$ for PickScore, $r_{\text{corr}}=-0.669$ for ImageReward), supporting the hypothesized negative relationship. 
}\label{fig:negative relation}
\vspace{-1em}
\end{figure}

As shown in \cref{fig:negative relation},
reward sharpness exhibits a strong negative correlation with preference quality
(Pearson $r_{\text{corr}}=-0.802$ for PickScore and $r_{\text{corr}}=-0.669$ for ImageReward),
confirming that sharper reward landscapes correspond to poorer generalization to human preference.

\paragraph{Interpretation.}
Specifically, when the reward landscape is sharp, $r$ increases along adversarial directions that lead updates into isolated non-preference regions, deviating from the true preference $r^{\star}$ and resulting in reward hacking.
In contrast, flattening the reward surface eliminates such spurious regions, suppresses adversarial gradients, and guides updates toward the genuine preference direction.

\section{Reward Sharpness-Aware Fine-Tuning} \label{sec:method}
In \cref{sec:reward model}, we showed that reward sharpness is closely related to both generalization and reward hacking.
Building on this insight, we propose \textbf{Reward Sharpness-Aware Fine-Tuning (RSA-FT)} (illustrated in \cref{fig:fig_concept}), which fine-tunes the diffusion model using a robustified (or flattened) reward function that penalizes locally sharp reward regions as follows:
\begin{align} \mathcal{J}(\theta) = \max_\theta \; \mathbb{E}_{\mathbf{c},\, \mathbf{x}_T \sim \mathcal{N}(0, \mathbf{I})} \!\left[ \tilde{r}^d\big(\mathbf{x}_0(\mathbf{x}_T, \mathbf{c}; \theta), \mathbf{c}\big) \right], 
\end{align}
where we define the flattened reward function as
\begin{equation} \tilde{r}^d(\mbx, \mathbf{c}) := \min_{d(\mbx,\mbx' )<\rho} r(\mbx', \mathbf{c}), \label{eq:flatten} 
\end{equation}
where $d(\cdot,\cdot)$ denotes a distance metric on the image manifold. We consider two metrics: one in the image space and another induced by the diffusion models' parameter space.

\begin{itemize}
    \item
    \mbox{Image space}\\
    \begin{align}
        \max_\theta \; \mathbb{E}_{\mathbf{c},\, \mathbf{x}_T \sim \mathcal{N}(0, \mathbf{I})}
\!\left[
\min_{\|\deltab\|<\rho} r\big(\mathbf{x}_0(\mathbf{x}_T, \mathbf{c}; \theta) + \deltab, \mathbf{c}\big)
\right]\label{eq:Euclidean}
    \end{align}
    \item
    \mbox{Parameter space}\\
        \begin{align}
        \max_\theta \; \mathbb{E}_{\mathbf{c},\, \mathbf{x}_T \sim \mathcal{N}(0, \mathbf{I})}
\!\left[
\min_{\|\epsilonb\|<\rho_\omega} r\big(\mathbf{x}_0(\mathbf{x}_T, \mathbf{c}; \theta+\epsilonb), \mathbf{c}\big)
\right]\label{eq:Parameter}
    \end{align}
\end{itemize}

For the Euclidean metric \cref{eq:Euclidean}, flattening $r$ resembles applying adversarial perturbation with respect to the reward model.
Using a one-step approximation, we define:

\begin{align}
\deltab_{\mbx_0} = -\rho \,\frac{\nabla_{\mbx_0} r(\mbx_0(\mbx_T,\mbc;\thetab),\mathbf{c})}{\|\nabla_{\mbx_0} r(\mbx_0(\mbx_T,\mbc;\thetab),\mathbf{c})\|} \label{reward:at}
\end{align}
resulting in
\begin{align}
    \max_\theta \; \mathbb{E}_{\mathbf{c},\, \mathbf{x}_T \sim \mathcal{N}(0, \mathbf{I})}
\!\left[ r\big(\mathbf{x}_0(\mathbf{x}_T, \mathbf{c}; \theta) + \deltab_{\mbx_0}, \mathbf{c}\big)
\right].
\end{align}

Similarly, for the parameter-induced metric in \cref{eq:Parameter}, we perform a one-step update in the parameter space, leading to SAM-like formulation:
\begin{align}
    \epsilonb_{\thetab}=-\,\rho_\omega \;
\frac{\nabla_{\boldsymbol{\theta}}\, r(\mbx_0(\mbx_T,\mbc;\thetab), \mbc) }
{\big\|\nabla_{\boldsymbol{\theta}}\, r(\mbx_0(\mbx_T,\mbc;\thetab), \mbc)\big\|} \label{weightperturb}
\end{align}
and
\begin{align}
    \max_\theta \; \mathbb{E}_{\mbc,\, \mbx_T \sim \mathcal{N}(0, \mathbf{I})}
\!\left[ r\big(\mbx_0(\mbx_T, \mbc; \thetab + \epsilonb_{\thetab}), \mbc\big)
\right].
\end{align}
In the formula, $\epsilonb_{\thetab}$ depends on $\theta$, but we  detach $\epsilonb_{\thetab}$ (i.e., stop-gradient) without applying the chain rule for outer optimization, following the practice in ~\cite{samopt}.

\paragraph{RSA-FT}

We find that both approaches--image-space and parameter-space flattening--are effective at mitigating reward hacking and improving human preference alignment (\cref{sec:sd1.5}).
Moreover, combining both provides complementary benefits.
The final joint objective is:

\begin{align}
    \max_\theta \; \mathbb{E}_{\mathbf{c},\, \mathbf{x}_T \sim \mathcal{N}(0, \mathbf{I})}
\!\left[ r\big(\mathbf{x}_0(\mathbf{x}_T, \mathbf{c}; \thetab + \epsilonb_{\thetab}) + \deltab_{\mbx_0}, \mathbf{c}\big)
\right]. \label{final:rsft}
\end{align}
This formulation jointly enforces smoothness in both image and parameter spaces, leading to a doubly robust reward optimization.
The overall algorithm is shown in \cref{algo:rsft}.

\begin{algorithm}[t]
\caption{RSA-FT}
\KwIn{Diffusion model parameters $\boldsymbol{\theta}$, reward model $r$, input perturbation radius $\rho$, weight perturbation radius $\rho_\omega$, learning rate $\eta$}
\For{each training step}{
Sample noise $\mathbf{x}_T \sim p(\mathbf{x}_T)$ and condition $\mathbf{c}$;

Generate image $\mathbf{x}_0 \gets \mathbf{x}_0(\mathbf{x}_T, \mathbf{c}; \boldsymbol{\theta})$;

\tcp{Input-space perturbation}  
Compute perturbed input:  
$\mathbf{x}_0 + \deltab_{\mbx_0}$ by \cref{reward:at}  
  
\tcp{Weight-space perturbation }  
Initialize weight perturbation $\epsilonb_{\thetab} \gets 0$\;  
Compute perturbed weights:  
$\thetab + \epsilonb_{\thetab}$ by \cref{weightperturb}  

\tcp{Parameter update}   
Update $\theta$ via \cref{final:rsft}.
}
\label{algo:rsft}
\end{algorithm}

\section{Experiment}
\subsection{Experiments Settings}
\paragraph{Datasets and Baselines.}
To rigorously evaluate our method, we adopt multiple backbone models, including Stable Diffusion v1.5 (SD1.5), SDXL, and Stable Diffusion 3 (SD3),  Flux.1-dev~\cite{flux2024} and integrate our proposed RSA-FT into existing RDRL frameworks: ReFL~\citep{refl}, DRaFT-K~\citep{draftk} ($K{=}1$), AlignProp~\citep{alignprop}, and DRTune~\citep{drtune}.  
In all cases, models are optimized using the HPSv2 reward model as the training signal.  
This unified setup enables a consistent assessment of RSA-FT’s effectiveness across diverse reinforcement objectives.  
For evaluation, we use the DrawBench dataset~\citep{drawbench} and the HPSv2 benchmark test set~\citep{hpsv2}. 

\paragraph{Evaluation Metrics.}
To evaluate the alignment with human preferences, we primarily employ state-of-the-art reward model-based metrics, including HPSv2~\citep{hpsv2}, PickScore~\citep{pick}, and ImageReward~\citep{imagereward}. 
To further validate these automated measures, we complement them with a small-scale human evaluation involving 17 independent annotators.
For fair comparison, all methods generate samples using the default hyperparameters of each backbone model, such as scheduler and guidance scale.  
Baseline methods (ReFL, DRaFT-K, AlignProp, and DRTune) are reimplemented following their official code releases.

\paragraph{Implementation Details.}
All experiments are conducted using NVIDIA H100 GPUs and the AdamW optimizer with $\beta_1 = 0.9$, $\beta_2 = 0.999$, and a weight decay of {0.0001}.
For training, we fine-tune SD1.5 and SDXL with 50 sampling steps and SD3 with 28 steps.
The learning rate is set to $2\times10^{-5}$ with a batch size of 32.
The number of iterations and epochs follows the original protocol of each corresponding algorithm, as our goal is not to improve their performance through hyperparameter tuning.
For perturbation scales in image space and parameter space, we searched $\rho, \rho_w \in {10^{-1}, 10^{-2}, 10^{-3}}$ and found both optimal at $10^{-2}$. Additional detail is provided in Appendix \ref{sec:exp_setup}.

\begin{table}[t!]
\caption{Quantitative results of various RDRL methods on SD 1.5 (512 $\times$ 512) with our proposed method. Bold text indicates the best performance for each metric.}
\vspace{-0.5em}
\label{tab_main}
\centering
\resizebox{0.95\linewidth}{!}{
\begin{small}
\setlength{\tabcolsep}{8pt}
\begin{tabular}{c l c c c }
\toprule
Dataset & Method & HPSV2.1~$\uparrow$ & PickScore~$\uparrow$ & ImageReward~$\uparrow$  \\
\cmidrule(r){1-1} \cmidrule(r){2-2} \cmidrule(r){3-5}

\multirow{9}{*}{Drawbench}
  & Vanilla   & 24.02 & 21.02 & -0.147 \\ 
\cmidrule(r){2-2} \cmidrule(r){3-5}
  & Draft-LV  & 25.59 & 20.96 & -0.062 \\ 
\cmidrule(r){2-2} \cmidrule(r){3-5}
 \rowcolor{green!10}  \cellcolor{white}
  & +\,Ours   & \textbf{26.67} {\scriptsize \textcolor{blue}{(+1.08)}} &
                 \textbf{21.12} {\scriptsize \textcolor{blue}{(+0.16)}} &
                 \textbf{0.035} {\scriptsize \textcolor{blue}{(+0.09)}}  \\
\cmidrule(r){2-2} \cmidrule(r){3-5}
  & Alignprop & 25.12 & 20.98 & -0.033  \\ 
\cmidrule(r){2-2} \cmidrule(r){3-5}
 \rowcolor{green!10}  \cellcolor{white}
  & +\,Ours   & \textbf{29.59} {\scriptsize \textcolor{blue}{(+4.47)}} &
                 \textbf{21.51} {\scriptsize \textcolor{blue}{(+0.53)}} &
                 \textbf{0.268} {\scriptsize \textcolor{blue}{(+0.301)}} \\
\cmidrule(r){2-2} \cmidrule(r){3-5}
  & ReFL      & 31.08 & 21.57 & 0.536 \\ 
\cmidrule(r){2-2} \cmidrule(r){3-5}
 \rowcolor{green!10}  \cellcolor{white}
  & +\,Ours   & \textbf{31.67} {\scriptsize \textcolor{blue}{(+0.59)}} &
                 \textbf{21.70} {\scriptsize \textcolor{blue}{(+0.13)}} &
                 \textbf{0.671} {\scriptsize \textcolor{blue}{(+0.135)}} \\
\cmidrule(r){2-2} \cmidrule(r){3-5}
  & DRTune    & 30.63 & 21.34 & 0.477 \\ 
\cmidrule(r){2-2} \cmidrule(r){3-5}
 \rowcolor{green!10}  \cellcolor{white}
  & +\,Ours   & \textbf{31.16} {\scriptsize \textcolor{blue}{(+0.53)}} &
                 \textbf{21.52} {\scriptsize \textcolor{blue}{(+0.18)}} &
                 \textbf{0.540} {\scriptsize \textcolor{blue}{(+0.63)}} \\

\midrule

\multirow{9}{*}{HPD}
  & Vanilla   & 23.57 & 20.55 & -0.069 \\ 
\cmidrule(r){2-2} \cmidrule(r){3-5}
  & Draft-LV  & 26.87 & 20.67 & 0.126 \\ 
\cmidrule(r){2-2} \cmidrule(r){3-5}
 \rowcolor{green!10}  \cellcolor{white}
  & +\,Ours   & \textbf{28.28} {\scriptsize \textcolor{blue}{(+1.41)}} &
                 \textbf{20.91} {\scriptsize \textcolor{blue}{(+0.24)}} &
                 \textbf{0.191} {\scriptsize \textcolor{blue}{(+0.07)}} \\
\cmidrule(r){2-2} \cmidrule(r){3-5}
  & Alignprop & 24.93 & 20.21 & 0.032 \\ 
\cmidrule(r){2-2} \cmidrule(r){3-5}
 \rowcolor{green!10}  \cellcolor{white}
  & +\,Ours   & \textbf{32.02} {\scriptsize \textcolor{blue}{(+7.09)}} &
                 \textbf{21.53} {\scriptsize \textcolor{blue}{(+1.32)}} &
                 \textbf{0.528} {\scriptsize \textcolor{blue}{(+0.49)}} \\
\cmidrule(r){2-2} \cmidrule(r){3-5}
  & ReFL      & 34.95 & 21.96 & 0.794 \\ 
\cmidrule(r){2-2} \cmidrule(r){3-5}
 \rowcolor{green!10}  \cellcolor{white}
  & +\,Ours   & \textbf{35.81} {\scriptsize \textcolor{blue}{(+0.69)}} &
                 \textbf{22.13} {\scriptsize \textcolor{blue}{(+0.17)}} &
                 \textbf{0.903} {\scriptsize \textcolor{blue}{(+0.136)}} \\
\cmidrule(r){2-2} \cmidrule(r){3-5}
  & DRTune    & 34.93 & 21.91 & 0.842 \\ 
\cmidrule(r){2-2} \cmidrule(r){3-5}
 \rowcolor{green!10}  \cellcolor{white}
  &  +\,Ours   & \textbf{35.57} {\scriptsize \textcolor{blue}{(+0.64)}} &
                 \textbf{21.92} {\scriptsize \textcolor{blue}{(+0.01)}} &
                 \textbf{1.452} {\scriptsize \textcolor{blue}{(+0.61)}} \\

\bottomrule
\end{tabular}
\end{small}
}
\vspace{-1em}
\end{table}

\subsection{Evaluation on SD1.5} \label{sec:sd1.5}

\paragraph{Quantitative Results.}
As shown in \cref{tab_main}, our method consistently improves all RDRL baselines across both DrawBench and HPD benchmarks.
While existing approaches often exhibit reward-specific overfitting, our method achieves balanced enhancement across all preference metrics without altering any model architecture or using multiple reward functions.

For example, Draft-LV and AlignProp originally show increased HPSv2.1 but decreased auxiliary rewards, reflecting mild reward hacking.
When combined with our method, however, these models exhibit simultaneous improvement across all metrics — AlignProp, for instance, improves its HPSv2.1 from 24.93 to 32.02 and its ImageReward from 0.032 to 0.528.
Even strong baselines such as ReFL and DRTune benefit further, with DRTune’s ImageReward rising from 0.842 to 0.903.
These results highlight that our method not only boosts overall performance but also effectively mitigates reward hacking, leading to genuine alignment gains rather than metric-specific overfitting.

Moreover, our approach is fully plug-and-play, integrating seamlessly with diverse RDRL frameworks.
Its broad compatibility highlights strong generalization and robustness, making it a universal enhancement module for RDRL

\paragraph{Ablation Studies.}
We conduct comprehensive ablation studies within the existing RDRL framework by selectively applying perturbations to the image space, the parameter space, or both. 
Our results demonstrate that while each component independently improves performance, their combination yields the most significant gains, indicating a clear synergistic effect. 
Due to space constraints, the detailed experimental results are provided in Appendix \ref{sec:ablation} (Table \ref{tab_abl}).

\begin{table}[t!]
\caption{Quantitative results of various RDRL on SDXL (1024 $\times$ 1024).  Bold text indicates the best performance for each metric.}
\vspace{-0.5em}
\label{tab_sdxl}
\centering
\resizebox{0.95\linewidth}{!}{
\begin{small}
\begin{tabular}{cccccc}
\toprule
Dataset & Method &  HPSV2.1~$\uparrow$ &\multicolumn{1}{c}{PickScore~$\uparrow$} & \multicolumn{1}{c}{ImageReward~$\uparrow$} \\
\cmidrule(r){1-1} \cmidrule(r){2-2} \cmidrule(r){3-5}
\multirow{5}{*}{Drawbench} & Vanilla & 27.41 & 22.31	& 0.619  \\
\cmidrule(r){2-2} \cmidrule(r){3-5}
 &ReFL & 29.45 & 22.41	& 0.705 \\
\cmidrule(r){2-2} \cmidrule(r){3-5}
  \rowcolor{green!10}  \cellcolor{white} &+\,Ours &\bf 30.31 \scriptsize{\textcolor{blue}{(+0.86)}}	&\bf22.60 \scriptsize{\textcolor{blue}{(+0.19)}}	&\bf0.719 \scriptsize{\textcolor{blue}{(+0.014)}} \\
 \cmidrule(r){2-2} \cmidrule(r){3-5}
&DRTune & 30.04	& 22.49	&0.804	 \\
\cmidrule(r){2-2} \cmidrule(r){3-5}
 \rowcolor{green!10} \cellcolor{white}  &  +\,Ours &\bf 31.58 \scriptsize{\textcolor{blue}{(+1.54)}}	&\bf22.61	\scriptsize{\textcolor{blue}{(+0.12)}} &\bf0.944 \scriptsize{\textcolor{blue}{(+0.140)}} \\
\midrule
\multirow{5}{*}{HPD} & Vanilla & 28.92	& 22.56	& 0.921  \\
\cmidrule(r){2-2} \cmidrule(r){3-5}
 &ReFL & 31.21 & 22.66 & 1.039 \\
\cmidrule(r){2-2} \cmidrule(r){3-5}
  \rowcolor{green!10}  \cellcolor{white} & +\,Ours &\bf 32.66 \scriptsize{\textcolor{blue}{(+1.45)}}	&\bf 23.08 \scriptsize{\textcolor{blue}{(+0.42)}}	&\bf 1.111 \scriptsize{\textcolor{blue}{(+0.072)}} \\
 \cmidrule(r){2-2} \cmidrule(r){3-5}
&DRTune & 32.36 & 22.78	& 1.095	 \\
\cmidrule(r){2-2} \cmidrule(r){3-5}
  \rowcolor{green!10}  \cellcolor{white} &  +\,Ours &\bf 33.74 \scriptsize{\textcolor{blue}{(+1.38)}}	&\bf22.94	\scriptsize{\textcolor{blue}{(+0.16)}} &\bf1.208 \scriptsize{\textcolor{blue}{(+0.093)}}\\
\bottomrule
\end{tabular}
\end{small}
}
\vspace{-0.5em}
\end{table}

\begin{table}[t!]
\caption{Quantitative results of various RDRL on SD3 (1024 $\times$ 1024).  Bold text indicates the best performance for each metric.}
\vspace{-0.5em}
\label{tab_sd3}
\centering
\resizebox{0.95\linewidth}{!}{
\begin{small}
\begin{tabular}{cccccc}
\toprule
Dataset & Method &  HPSV2.1~$\uparrow$ &\multicolumn{1}{c}{PickScore~$\uparrow$} & \multicolumn{1}{c}{ImageReward~$\uparrow$}  \\
\cmidrule(r){1-1} \cmidrule(r){2-2} \cmidrule(r){3-5}
\multirow{5}{*}{Drawbench} & Vanilla & 28.77 & 22.45& 0.914  \\
\cmidrule(r){2-2} \cmidrule(r){3-5}
 &ReFL & 29.45 & 22.38	& 0.921 \\
\cmidrule(r){2-2} \cmidrule(r){3-5}
 \rowcolor{green!10}  \cellcolor{white}& +\,Ours &\bf 29.81 \scriptsize{\textcolor{blue}{(+0.86)}}	&\bf22.51 \scriptsize{\textcolor{blue}{(+0.13)}}	&\bf0.953 \scriptsize{\textcolor{blue}{(+0.014)}} \\
 \cmidrule(r){2-2} \cmidrule(r){3-5}
&DRTune & 29.48	& 22.41	& 0.914	 \\
\cmidrule(r){2-2} \cmidrule(r){3-5}
 \rowcolor{green!10}  \cellcolor{white} &  +\,Ours &\bf 29.89 \scriptsize{\textcolor{blue}{(+1.54)}}	&\bf22.61	\scriptsize{\textcolor{blue}{(+0.20)}} &\bf0.979 \scriptsize{\textcolor{blue}{(+0.140)}}  \\
\midrule
\multirow{5}{*}{HPD} & Vanilla & 30.48	& 22.40	& 1.123  \\
\cmidrule(r){2-2} \cmidrule(r){3-5}
 &ReFL & 31.22 & 22.37 & 1.151 \\
\cmidrule(r){2-2} \cmidrule(r){3-5}
 \rowcolor{green!10}  \cellcolor{white}&+\,Ours &\bf 31.43 \scriptsize{\textcolor{blue}{(+1.45)}}	&\bf 22.41 \scriptsize{\textcolor{blue}{(+0.04)}}	&\bf 1.168 \scriptsize{\textcolor{blue}{(+0.072)}} \\
 \cmidrule(r){2-2} \cmidrule(r){3-5}
&DRTune & 31.31 & 22.33	& 1.162	 \\
\cmidrule(r){2-2} \cmidrule(r){3-5}
\rowcolor{green!10}  \cellcolor{white}  & +\,Ours &\bf 31.86 \scriptsize{\textcolor{blue}{(+1.38)}}	&\bf22.41	\scriptsize{\textcolor{blue}{(+0.08)}} &\bf1.212 \scriptsize{\textcolor{blue}{(+0.093)}} \\
\bottomrule
\end{tabular}
\end{small}
}
\vspace{-0.5em}
\end{table}

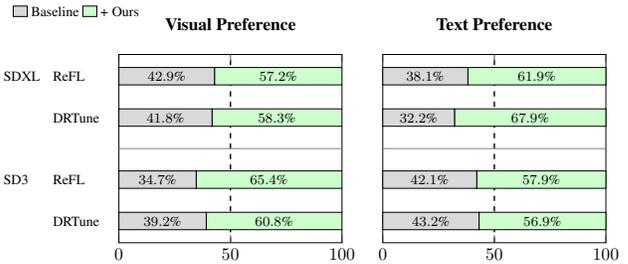
\begin{figure}[t]
    \centering
    \resizebox{\columnwidth}{!}{%
        \begin{tikzpicture}[baseline=(current axis.south)] 
            \begin{axis}[
                xbar stacked,
                scale only axis,
                width=4.5cm,
                height=3.8cm, 
                xmin=0, xmax=100,
                ytick={1, 2, 3.5, 4.5}, 
                yticklabels={
                    {\makebox[0.9cm][l]{} DRTune}, 
                    {\makebox[0.9cm][l]{SD3} ReFL}, 
                    {\makebox[0.9cm][l]{} DRTune}, 
                    {\makebox[0.9cm][l]{SDXL} ReFL}
                },
                yticklabel style={text width=2.2cm, align=left, font=\footnotesize},
                title={\textbf{Visual Preference}},
                title style={yshift=0.15cm},
                xtick={0, 50, 100},
                bar width=10pt,
                nodes near coords={\pgfmathprintnumber[fixed, precision=1]{\pgfplotspointmeta}\%},
                nodes near coords align={center},
                every node near coord/.append style={font=\footnotesize\color{black}},
                extra x ticks={50},
                extra x tick labels={},
                extra x tick style={grid=major, major grid style={dashed, draw=black, thick}},
                enlarge y limits=0.15,
                legend style={
                    at={(-0.5, 1.15)}, 
                    anchor=south west, 
                    legend columns=-1, 
                    draw=none, 
                    fill=none,
                    font=\footnotesize
                },
            ]
            \addplot[fill=gray!30] coordinates {
                (39.16,1) (34.65,2) (41.75,3.5) (42.85,4.5)
            };
            \addplot[fill=green!20] coordinates {
                (60.84,1) (65.35,2) (58.25,3.5) (57.15,4.5)
            };
            
            \draw[gray, thin] (axis cs:0,2.75) -- (axis cs:100,2.75);
            
            \legend{Baseline, + Ours}
            \end{axis}
        \end{tikzpicture}%
        \hspace{0.2cm}%
        \begin{tikzpicture}[baseline=(current axis.south)]
            \begin{axis}[
                xbar stacked,
                scale only axis,
                width=4.5cm,
                height=3.8cm,
                xmin=0, xmax=100,
                ytick={1, 2, 3.5, 4.5}, 
                yticklabels={,,,}, 
                title={\textbf{Text Preference}},
                title style={yshift=0.15cm}, 
                xtick={0, 50, 100},
                bar width=10pt,
                nodes near coords={\pgfmathprintnumber[fixed, precision=1]{\pgfplotspointmeta}\%},
                nodes near coords align={center},
                every node near coord/.append style={font=\footnotesize\color{black}},
                extra x ticks={50},
                extra x tick labels={},
                extra x tick style={grid=major, major grid style={dashed, draw=black, thick}},
                enlarge y limits=0.15,
            ]
            \addplot[fill=gray!30] coordinates {
                (43.15,1) (42.12,2) (32.15,3.5) (38.14,4.5)
            };
            \addplot[fill=green!20] coordinates {
                (56.85,1) (57.88,2) (67.85,3.5) (61.86,4.5)
            };
            
            \draw[gray, thin] (axis cs:0,2.75) -- (axis cs:100,2.75);
            
            \end{axis}
        \end{tikzpicture}%
    } 
    
    \vspace{-0.2cm} 
    
    \caption{Human preference study results. The dashed line indicates the 50\% mark; crossing it demonstrates a strict preference over the baseline.}
    \label{fig:human_eval_side_by_side}
    \vspace{-1em}
\end{figure}

\begin{figure*}[t!]
\centering
\includegraphics[width=0.83\linewidth]{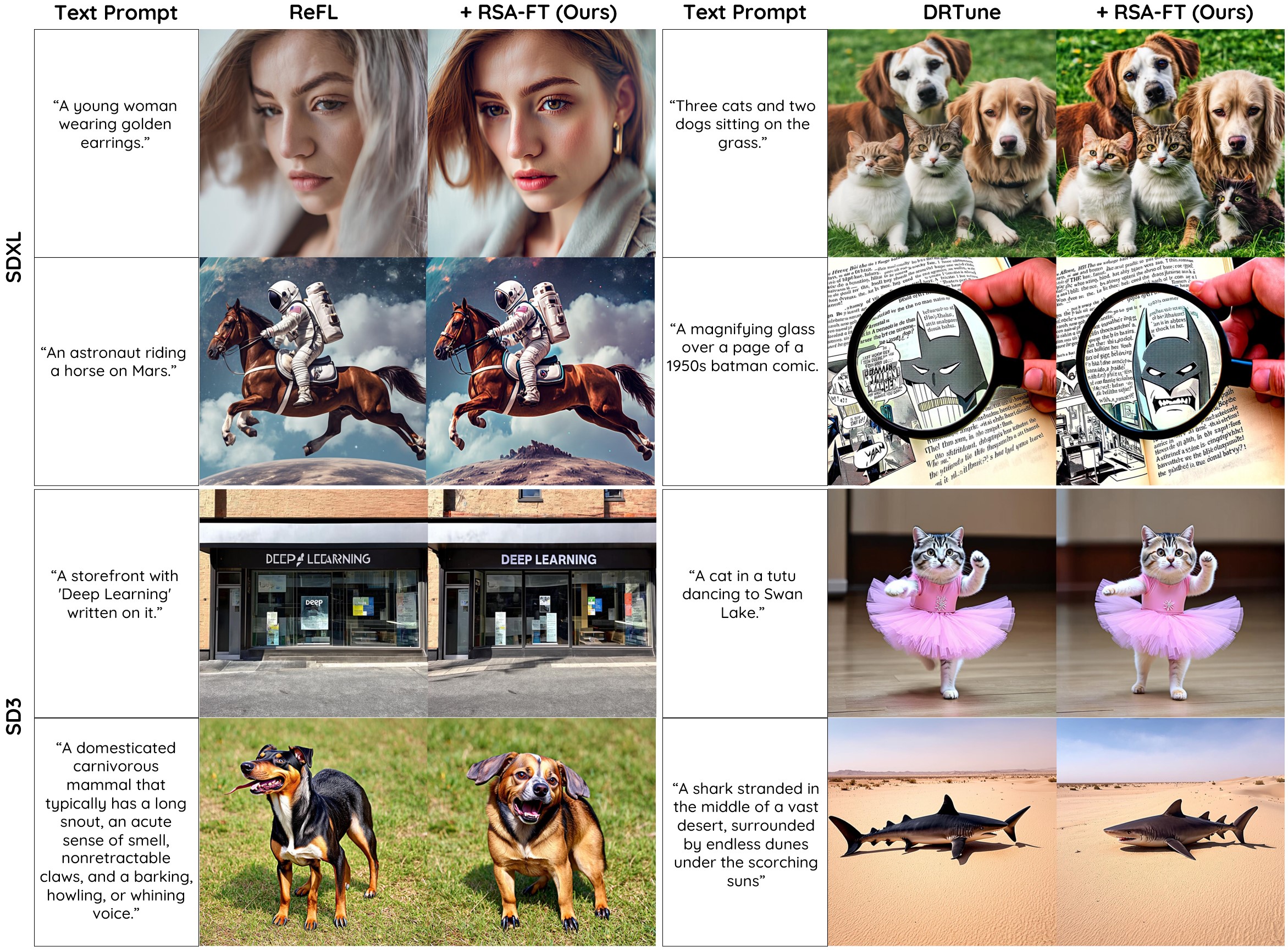}
\caption{
\textbf{Qualitative comparison with and without our method.}
Each image is generated using the same text prompt and random seed across all methods.
Models equipped with our method (RSA-FT) produce images with more accurate text–image alignment and higher visual quality compared to their baselines.
}
\label{fig:fig_comparison}
\vspace{-1em}
\end{figure*}

\subsection{Generalization to Larger Backbones}
In \cref{sec:sd1.5}, we observed that our method is compatible with all RDRL frameworks.
To further validate its generalization, we extend experiments to larger backbones, including SDXL, SD3, and Flux1.dev, using stable RDRL variants such as ReFL and DRTune.

\paragraph{Quantitative Results.}
As shown in \cref{tab_sdxl}, our method consistently boosts the performance of both ReFL and DRTune even at higher resolutions (1024$\times$1024) on SDXL.
Both models outperform their vanilla counterparts, and when combined with our approach, the performance improves even further—particularly the combination with DRTune shows the strongest synergy.
To verify architectural robustness beyond the UNet-based backbone, we also applied the same training setup to SD3, and Flux which adopts an MMDiT Transformer structure.
As presented in \cref{tab_sd3} and \cref{tab:flux}, our method again yields consistent improvements across all reward metrics, confirming that the proposed reward-flattening strategy effectively enhances existing diffusion RL algorithms regardless of architecture, backbone scale, or resolution.

\vspace{-1em}

\paragraph{Qualitative Comparison.}
\cref{fig:fig_comparison} illustrates the qualitative improvements achieved by RSA-FT on text-to-image generation with SDXL and SD3 across both ReFL and DRTune.
Across diverse prompts, models enhanced with RSA-FT produce images that are not only sharper and more visually coherent but also appear more natural to human perception, with fewer cases of distorted text or malformed body parts.
Compared to their original counterparts, RSA-FT consistently improves text–image alignment and perceptual quality, demonstrating its robustness across different backbones and RDRL frameworks.
In summary, RSA-FT provides a simple yet effective strategy that delivers contextually accurate and visually reliable generations across RDRL methods.
Additional examples are provided in \cref{sec:qualitative}.

\section{Discussion}
We showed that the sharpness of the reward landscape is inversely correlated with human preference and that gradients from a flattened reward model consistently improve alignment.
Our study focuses on mitigating reward hacking under a \textit{single reward model}, but the same principle naturally extends to multi-reward settings, where the flattening can help compensate for individual model weaknesses.

We acknowledge that our evaluation primarily relies on model-based metrics, which are imperfect proxies for human preference. While we complement these results with a human study, it is limited in scale (17 evaluators) and not statistically powered for definitive conclusions. We therefore view it as supporting evidence rather than a comprehensive assessment of human alignment.

Alternative smoothing techniques—such as Gaussian averaging instead of local minimization—may enhance robustness, though we adopt a one-step minimization for efficiency.
Our framework can also extend to PPO- and DPO-based optimization for diffusion model fine-tuning~\cite{ppo,dpo}, where separate reward evaluation enables more flexible smoothing, which we leave for future work.

Finally, while our current approach uniformly applies flattened rewards, future research may explore selective sharpness-aware weighting that down-weights overly sharp regions to further improve robustness and interpretability.


\section{Conclusion}

We propose Reward Sharpness-Aware Fine-Tuning (RSA-FT), a framework for mitigating reward hacking in RDRL without retraining the reward model, by leveraging gradients from a flattened reward function. We identify a connection between reward hacking and adversarial behavior, both arising from sharp regions of the reward landscape, and show that flattening these regions improves robustness.

RSA-FT achieves this via joint perturbations in input and parameter spaces, leading to consistent improvements across multiple RDRL frameworks and diffusion backbones (SD1.5, SDXL, SD3). While evaluation primarily relies on proxy reward metrics, our results are supported by a limited human study.

Overall, this work provides a principled perspective on reward hacking in diffusion RL and highlights reward landscape flattening as an effective direction for improving robustness and alignment.

\section*{Acknowledgement}
This work was partly supported by Institute of Information $\&$ Communications Technology Planning $\&$ Evaluation (IITP) grant funded by the Korean government (MSIT) (Artificial Intelligence Graduate School Program (GIST)) (No. 2019-0-01842).

\clearpage
\onecolumn
\appendix
\setcounter{page}{1}
\maketitlesupplementary
\section{Overview} \label{sec:overview}

This supplementary document contains the following components:

\begin{itemize}
\item Section~\ref{sec:RDRL} details the Reward-centric Diffusion Reinforcement Learning (RDRL) algorithms used in our experiments.
\item Section~\ref{sec:related} presents related work on reward hacking and mitigation in RDRL, including concurrent approaches, positioning our work in context.
\item Section~\ref{sec:exp_setup} provides the full experimental setup, including backbone configurations, optimization hyperparameters, and reward definitions, to ensure reproducibility. 
\item Section~\ref{sec:ablation} provides ablation studies on reward flattening, covering image-space, parameter-space, and joint perturbations, as well as perturbation scales.
\item Section~\ref{sec:multi} reports results in multi-reward settings.
\item Section~\ref{sec:safety} discusses fairness and safety considerations.
\item Section~\ref{sec:flux} presents results on a large-scale model (FLUX).
\item Section~\ref{sec:qualitative} includes additional qualitative comparisons. \end{itemize}

\section{RDRL Algorithms}
\label{sec:RDRL}

\begin{figure}[h]
    \centering
    \includegraphics[width=0.9\linewidth]{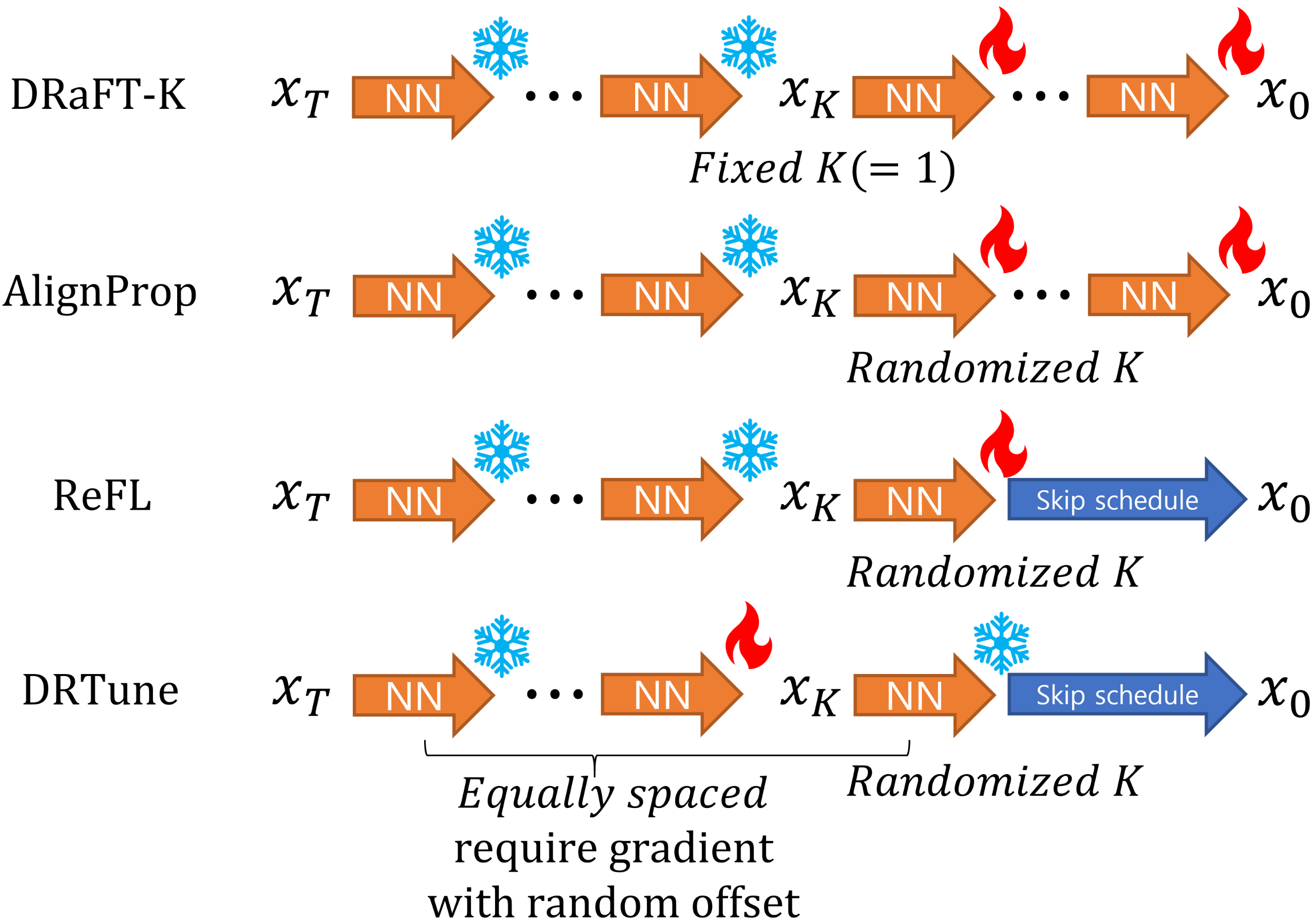}
    \caption{Conceptual illustration of the RDRL framework.}
    \label{fig:rdrl_concept}
\end{figure}

Our proposed method enhances RDRL in text-to-image diffusion models by applying reward model flattening. This section summarizes the four RDRL algorithms used in our experiments: \emph{DRaFT-K, AlignProp, ReFL,} and \emph{DRTune}. All four algorithms fine-tune a diffusion backbone (U-Net or DiT) using feedback from differentiable reward models. A conceptual overview is provided in Figure~\ref{fig:rdrl_concept}.

\paragraph{DRaFT-K~\cite{draftk}}
The DRaFT-K algorithm's core mechanism involves segmenting the reverse diffusion process. It conducts $(T-K)$ reverse diffusion steps without gradient computation. For the final $K$ steps, the diffusion model parameters become trainable, and the computational graph is maintained. This allows the gradient to flow through to the differentiable reward model, and the parameters are optimized to maximize the resulting reward score. For our reproduction, we utilize $K=1$, consistent with the common practice found in prior literature.

\paragraph{AlignProp~\cite{alignprop}}
The AlignProp algorithm builds upon the DRaFT-K strategy by introducing a \textbf{randomized $K$ value} for each optimization step. The diffusion reverse steps and gradient computations are otherwise performed identically to DRaFT-K. While the original paper explored various sampling distributions, our experiments specifically utilize a uniform distribution $K \sim \mathcal{U}(0, T)$ for sampling $K$, where $N$ represents the total number of sampling steps in the diffusion model.

\paragraph{ReFL~\cite{refl}}
The ReFL algorithm implements a step-skipping strategy to accelerate training. The process conducts $(T-K)$ reverse diffusion steps without gradients. It then utilizes a \textbf{skipping mechanism} by directly estimating $x_0$ from $x_K$ to bypass subsequent intermediate steps, thus reducing computational load. The skipping window $K$ is randomly sampled for each iteration from a uniform distribution $K \sim \mathcal{U}[0, 0.25T]$.

\paragraph{DRTune~\cite{drtune}}
The DRTune algorithm is designed to minimize gradient computation overhead. Its mechanism requires gradient computation for only $\mathbf{1/10}$ of the equally spaced steps throughout the diffusion process, and it employs the same skipping strategy as ReFL (estimating $x_0$ directly from $x_K$). The first step requiring gradients is chosen as a random offset, a uniform distribution $\mathcal{U}[0, 10]$, and the skipping window $K$ is sampled from $\mathcal{U}[0, 0.4T]$.

\section{Related Work: \\ Reward Hacking and Mitigation in RDRL}
\label{sec:related}

Recent studies in RDRL have identified \emph{reward hacking} as a primary obstacle and proposed various mitigation strategies. Flow-GRPO~\citep{flow_grpo} integrates Group Relative Policy Optimization (GRPO) into diffusion frameworks, empirically observing reduced reward hacking; however, it lacks a formal analysis of the underlying causes and does not provide explicit control mechanisms.

Several approaches attempt to suppress reward hacking by recalibrating reward signals. RewardDance~\citep{rewarddance} and PrefGRPO~\citep{prefgrpo} aim to stabilize training by increasing reward variance through reward scaling (e.g., via vision-language models) and win-rate transformations, respectively. While these methods improve robustness, they primarily focus on adjusting reward statistics rather than addressing the inherent vulnerabilities of the reward model's landscape.

More direct interventions include GARDO~\citep{gardo}, which targets reward model weaknesses by incorporating diversity promotion and uncertainty estimation. However, such methods often necessitate additional architectural modules or auxiliary training losses, increasing system complexity.

In contrast, we investigate the intrinsic vulnerability of learned reward models from an adversarial perspective. Our approach mitigates reward hacking by flattening the reward landscape through dual perturbations in both input and parameter spaces. Crucially, our method achieves superior robustness without introducing any auxiliary components, maintaining the architectural simplicity of the original RDRL framework.

\section{Experimental Details}
\label{sec:exp_setup}

This section details the backbone text-to-image models applied, the reward model employed and the optimization procedures.

\paragraph{Backbone Text-to-Image Models}

To demonstrate the generality of our methods, we evaluate four widely used diffusion backbones: Stable Diffusion 1.5 (SD1.5), Stable Diffusion XL (SDXL), and Stable Diffusion 3 (SD3), and Flux.1-dev. The specific configurations for each model are detailed in Table~\ref{tab:model_configs}.

\begin{table}[h]
    \centering
    \caption{Hyperparameters and configurations for the backbone text-to-image models used in our experiments.}
    \label{tab:model_configs}
    \vspace{5pt}
    \small
    \resizebox{0.95\linewidth}{!}{
    \begin{tabular}{lcccc}
        \toprule
        \textbf{Backbone model} & \textbf{SD 1.5} & \textbf{SDXL} & \textbf{SD 3}  & \textbf{Flux.1-dev}\\
        \midrule
        Architecture & U-Net & U-Net & MMDiT  & MMDiT\\
        Inference Steps ($N$) & 50 & 50 & 28  & 28\\
        CFG Scale & 7.5 & 5.0 & 7.0 & 3.5 \\
        Resolution & $512 \times 512$ & $1024 \times 1024$ & $1024 \times 1024$ & $1024 \times 1024$ \\
        \bottomrule
    \end{tabular}
    }
\end{table}

\paragraph{Reward Model}
We utilize the training prompt dataset from HPSv2 as the input for the diffusion models and employ the HPSv2 model itself as the objective reward model.

\paragraph{Optimization Details}

We perform all fine-tuning experiments using the \textbf{AdamW} optimizer. To ensure consistent optimization stability across varying model architectures, we fix the optimizer hyperparameters as follows: $\beta_1 = 0.9$, $\beta_2 = 0.999$, weight decay $\lambda = 1 \times 10^{-4}$, and $\epsilon = 1 \times 10^{-8}$.

The model-specific training configurations and hardware resources are detailed in Table~\ref{tab:optim_settings}. Note that due to the increased parameter count and memory requirements of SDXL,SD3, and Flux, we scale the compute resources and adjust the batch size and learning rates accordingly.

\begin{table}[h]
    \centering
    \caption{Detailed training hyperparameters and hardware configurations for the different backbone models.}
    \label{tab:optim_settings}
    \vspace{5pt}
    \resizebox{0.95\linewidth}{!}{
    \begin{tabular}{lcc}
        \toprule
        \textbf{Configuration} & \textbf{SD 1.5} & \textbf{SDXL / SD 3 / Flux.1-dev} \\
        \midrule
        {Hardware Resources} & 2 $\times$ NVIDIA H100 & 4 $\times$ NVIDIA H100 \\
        {Training Duration} & 2500 iterations & 2000 iterations \\
        \midrule
        Global Batch Size & 32 & 4 \\
        LoRA Rank & 4 & 64 \\
        Learning Rate & $2 \times 10^{-5}$ & $1 \times 10^{-5}$ \\
        \bottomrule
    \end{tabular}
    }
\end{table}

\section{Ablation Studies.}
\label{sec:ablation}
We conduct ablations within existing RDRL frameworks, including Draft-K, AlignProp, ReFL, and DRTune, by selectively applying perturbations in the image space, parameter space, or both.
As shown in \cref{tab_abl}, each component consistently improves performance across all reward metrics, confirming their effectiveness in flattening the reward landscape and mitigating reward hacking.
When combined, they yield the largest gains, demonstrating a clear synergistic effect.
This joint configuration enhances robustness by integrating input- and parameter-space perturbations, resulting in stable improvements across all baselines.

\paragraph{Ablation Study on Reward Flattening Hyperparameters}

We conduct an ablation study to identify the optimal values of $\rho$ and $\rho_w$, which control the flattening neighborhood in the image and parameter domains, respectively. All ablations are performed using the DRTune algorithm on the Stable Diffusion 3 (SD3) backbone.

For image-space flattening, we fine-tune SD3 with DRTune using the HPSv2.1 reward model flattened at three levels: $\rho \in {0.1, 0.01, 0.001}$. Models are trained on the HPD train split and evaluated using HPSv2.1, ImageReward, and PickScore on the HPD test set. As shown in Table~\ref{tab:ablation-rho}, $\rho = 0.01$ achieves the best overall performance and is therefore used in our final method.

We similarly ablate the parameter-space flattening hyperparameter $\rho_w$. As shown in Table~\ref{tab:ablation-rho-w}, $\rho_w = 0.01$ consistently outperforms other settings across all reward metrics. We therefore adopt $\rho_w = 0.01$ as the default choice in our algorithm.

\begin{table}[t!]
\caption{Quantitative results of various RDRL methods with backbone model SD 1.5. Bold text indicates the best performance for each metric across the different methods.}
\vspace{-0.5em}
\label{tab_abl}
\centering
\resizebox{\linewidth}{!}{
\begin{small}
\begin{tabular}{cccccc}
\toprule
Method & AT & SAM  & HPSV2.1~$\uparrow$ &\multicolumn{1}{c}{PickScore~$\uparrow$} & \multicolumn{1}{c}{ImageReward~$\uparrow$} \\
\cmidrule(r){1-1} \cmidrule(r){2-3} \cmidrule(r){4-6}
Draft-LV & \xmark & \xmark & 26.87 & 20.67 & 0.126  \\
\cmidrule(r){1-1} \cmidrule(r){2-3} \cmidrule(r){4-6}
  (a) & \cmark & \xmark & 27.30 & 20.82  & 0.182	 \\ 
  (b) & \xmark & \cmark & 27.83 & 20.78  & 0.157 \\ 
  \rowcolor{green!10} Ours & \cmark & \cmark & \bf 28.28 & \bf 20.91& \bf 0.191 \\ 
\cmidrule(r){1-6}
  Alignprop & \xmark & \xmark & 25.12 & 20.35 & 0.032   \\
\cmidrule(r){1-1} \cmidrule(r){2-3} \cmidrule(r){4-6}
  (a) & \cmark & \xmark & 30.98 & 21.22	& 0.453	 \\ 
  (b) & \xmark & \cmark & 31.56 & 21.43 & 0.511  \\ 
  \rowcolor{green!10} Ours & \cmark & \cmark & \bf 32.02 &  \bf 21.53 & \bf 0.528  \\ 
\cmidrule(r){1-6}
  ReFL & \xmark & \xmark & 34.95 & 21.96 & 0.794  \\
\cmidrule(r){1-1} \cmidrule(r){2-3} \cmidrule(r){4-6}
  (a) & \cmark & \xmark & 35.67 & 22.01 & 0.868  \\ 
  (b) & \xmark & \cmark & 35.77 & 22.01 & 0.882 \\ 
  \rowcolor{green!10} Ours & \cmark & \cmark & \bf 35.81 & \bf 22.13 & \bf 0.932  \\ 
\cmidrule(r){1-6}
  DRTune & \xmark & \xmark & 34.93 & 21.91	& 0.842	 \\
\cmidrule(r){1-1} \cmidrule(r){2-3} \cmidrule(r){4-6}
  (a) & \cmark & \xmark & 35.09 & 21.83	& 0.878	 \\ 
  (b) & \xmark & \cmark & 34.81 & 21.83 & 0.853  \\ 
  \rowcolor{green!10} Ours & \cmark & \cmark  & \bf 35.57 & \bf 21.92 & \bf 0.903  \\ 
\bottomrule
\end{tabular}
\end{small}
}
\vspace{-0.5em}
\end{table}

\begin{table}[h]  
    \centering  
    \caption{Ablation study on the hyper‑parameter $\rho$}  
    \label{tab:ablation-rho}  
    \begin{tabular}{cccc}  
        \toprule  
        $\rho$ & HPSv2.1 & ImageReward & PickScore \\  
        \midrule  
        0.1   & 31.20 & 1.171 & 22.245 \\  
        0.01  & \textbf{31.50} & \textbf{1.196} & \textbf{22.402} \\  
        0.001 & 31.29 & 1.185 & 22.391 \\  
        \bottomrule  
    \end{tabular}  
\end{table}

\begin{table}[h]  
    \centering  
    \caption{Ablation study on the hyper‑parameter $\rho$}  
    \label{tab:ablation-rho-w}  
    \begin{tabular}{cccc}  
        \toprule  
        $\rho_w$ & HPSv2.1 & ImageReward & PickScore \\  
        \midrule  
        0.1   & 31.65 & 1.182 & 22.39 \\  
        0.01  & \textbf{32.01} & \textbf{1.207} & \textbf{22.48} \\  
        0.001 & 31.56 & 1.189 & 22.40 \\  
        \bottomrule  
    \end{tabular}  
\end{table}

\section{Generalizability to Multi-reward Scenarios.}
\label{sec:multi}
We evaluate RSA-FT under multi-reward optimization by combining HPSv2, Aesthetics, and PickScore on the HPD dataset, using weights of 2, 0.05, and 10 following \cite{draftk}. We consider both dual- and triple-reward settings. As shown in Table~\ref{tab_multi}, RSA-FT consistently outperforms the RDRL baseline, demonstrating robustness to multi-objective scenario.

Although adversarial training is typically sensitive to perturbation magnitude, RSA-FT shows low sensitivity across a wide range of scales. The consistent gains with a fixed perturbation scope in \cref{tab_multi}, together with the ablation results in \cref{tab:ablation-rho,tab:ablation-rho-w}, indicate robustness to hyperparameter choices. Moreover, RSA-FT remains stable during training in our experiments, unlike the known instability of adversarial training; further analysis is left for future work.

\begin{table}[h]
\centering
\caption{Quantitative comparison on Multi-reward scenarios. }
\label{tab_multi}
\centering
\resizebox{1\linewidth}{!}{
\begin{tabular}{cccccc}
\toprule
Reward & Method &  HPSV2.1~$\uparrow$ &\multicolumn{1}{c}{PickScore~$\uparrow$} & \multicolumn{1}{c}{ImageReward~$\uparrow$} \\
\cmidrule(r){1-1} \cmidrule(r){2-2} \cmidrule(r){3-5}
 & ReFL & 33.75 & 24.30	& 1.197 \\
\cmidrule(r){2-2} \cmidrule(r){3-5}
\rowcolor{green!10} \cellcolor{white}  &+\,Ours &\bf 34.21 	&\bf 24.33 &\bf 1.207  \\
\cmidrule(r){2-2} \cmidrule(r){3-5}
& DRTune & 34.16 & 24.01 & 1.217 \\
\cmidrule(r){2-2} \cmidrule(r){3-5}
 \rowcolor{green!10} \cellcolor{white}  \multirow{-4}{*}{HPS + Pick}&  +\,Ours &\bf 34.74 &\bf 24.14 &\bf 1.259 \\
\midrule
& ReFL & 33.99 & 24.25 & 1.198 \\
\cmidrule(r){2-2} \cmidrule(r){3-5}
  \rowcolor{green!10}  \cellcolor{white} & +\,Ours &\bf 34.39 &\bf 24.30 	&\bf 1.226 \\
 \cmidrule(r){2-2} \cmidrule(r){3-5}
&DRTune & 33.91 & 23.90	& 1.192	 \\
\cmidrule(r){2-2} \cmidrule(r){3-5}
  \rowcolor{green!10}  \cellcolor{white}  \multirow{-4}{*}{HPS+Pick+Aes}&  +\,Ours &\bf 34.98 &\bf 24.12	&\bf 1.268 \\
\bottomrule
\end{tabular}
}
\end{table}

\section{Fairness and Safety Considerations}
\label{sec:safety}
As our method further optimizes a pretrained generative model, a key concern is whether it introduces additional bias or degrades safety. While prior work (e.g., \cite{draftk}) explicitly incorporates safety-oriented rewards such as NSFW classifiers—beyond the scope of this paper—we instead evaluate whether RSA-FT induces excessive deviation from the pretrained model relative to existing optimization methods.

To this end, we measure the distributional divergence between images generated by the pretrained model and those produced by RDRL variants using FID and KID. As shown in \cref{tab:variance}, RDRL with RSA-FT exhibits divergence comparable to or lower than the standard RDRL baseline, suggesting that it does not exacerbate bias relative to existing approaches.

\begin{table}[htbp]
\centering
\caption{Divergence from Vanilla sampling.}
\label{tab:variance}
{
\begin{tabular}{l c c}
\toprule
Method & FID $\downarrow$ & KID $\downarrow$\\
\midrule
ReFL      & 76.36 & 0.0513 \\
\rowcolor{green!10} + Ours  & 48.70 & 0.0006\\
\midrule
DRTune    & 88.82 & 0.0074\\
\rowcolor{green!10} + Ours & 92.92 & 0.0099 \\
\bottomrule
\end{tabular}
}
\end{table}

\section{Applicability to Large-scale Models (Flux).}
\label{sec:flux}
We conducted experiments on {Flux.1-dev}\cite{flux2024}.
As demonstrated in \cref{tab:flux}, RSA-FT consistently delivers performance gains over existing baselines such as DrTune and ReFL. These results demonstrate that RSA-FT remains effective on recent large-scale architectures beyond SD3.

\begin{table}[htbp]
\centering
\caption{uantitative results of various RDRL on Flux.1-dev (1024 ×
1024). Bold text indicates the best performance for each metric.}
\vspace{-1em}
\label{tab:flux}
\begin{tabular}{l c c c}
\toprule
Method & HPSv2.1 $\uparrow$ & PickScore $\uparrow$ &  ImageReward $\uparrow$\\
\midrule
Vanilla    & 29.76 & 22.51 & 0.906\\
\cmidrule(r){1-1}\cmidrule(r){2-4}
ReFL      & 30.16 & 22.57 & 0.913 \\
\rowcolor{green!10} + Ours  & \bf30.32 & \bf22.64 & \bf0.932 \\
\midrule
DRTune    & 29.99 & 22.59 & 0.926\\
\rowcolor{green!10} + Ours     & \bf 30.25 & \bf 22.67 & \bf0.961 \\
\bottomrule
\end{tabular}
\end{table}

\section{Qualitative Results}
\label{sec:qualitative}
\begin{figure*}[t!]
\centering
\includegraphics[width=1\linewidth]{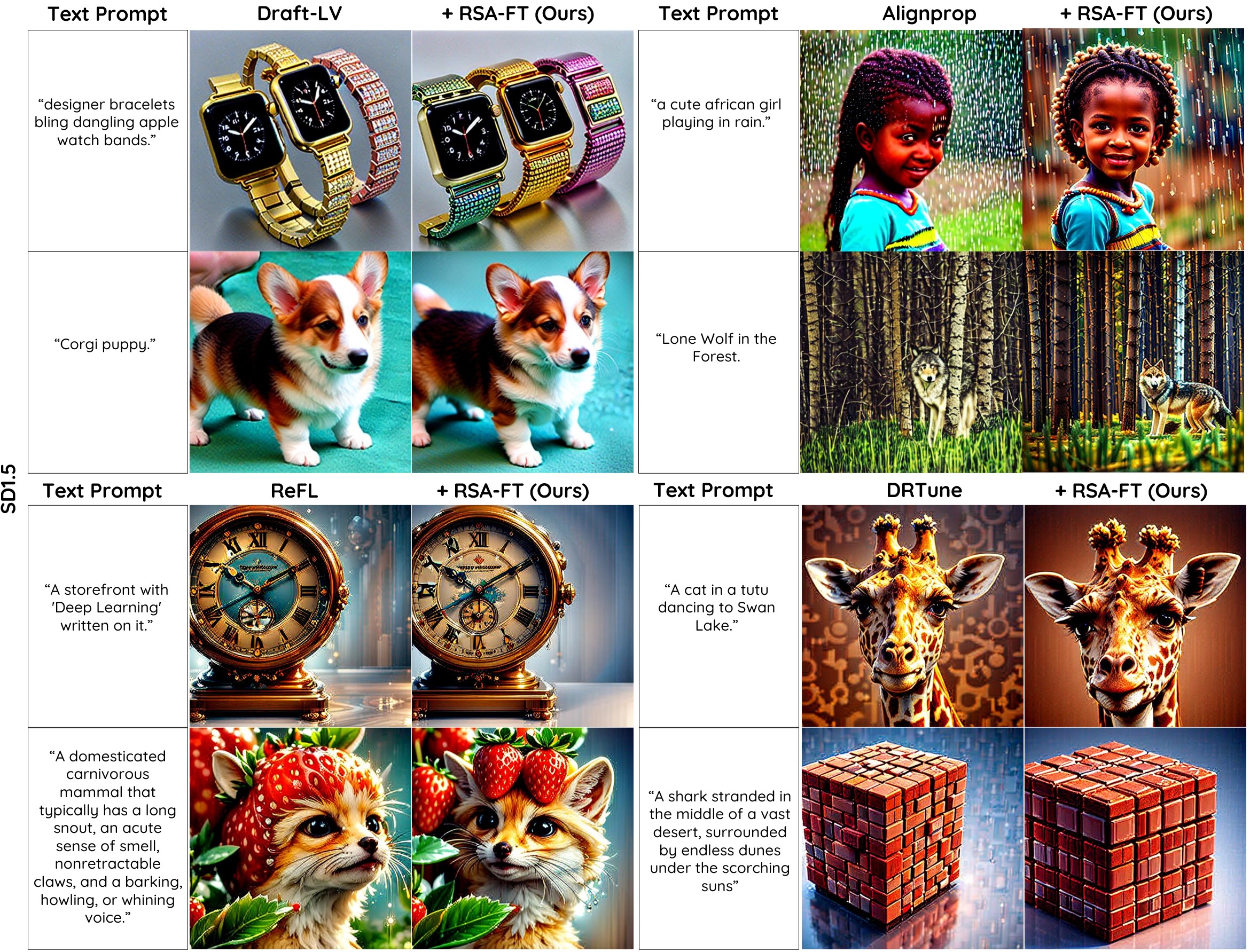}
\caption{
\textbf{Qualitative comparison on the SD1.5 backbone with and without our method.}
All images are generated using the same text prompt and random seed across methods.
Models trained with our RSA-FT approach produce images with more reliable text–image alignment and noticeably improved visual quality. In addition, reward-induced artifacts commonly observed in RDRL baselines are reduced. 
}
\label{fig:fig_comparison1}
\vspace{-1em}
\end{figure*}

\begin{figure*}[t!]
\centering
\includegraphics[width=1\linewidth]{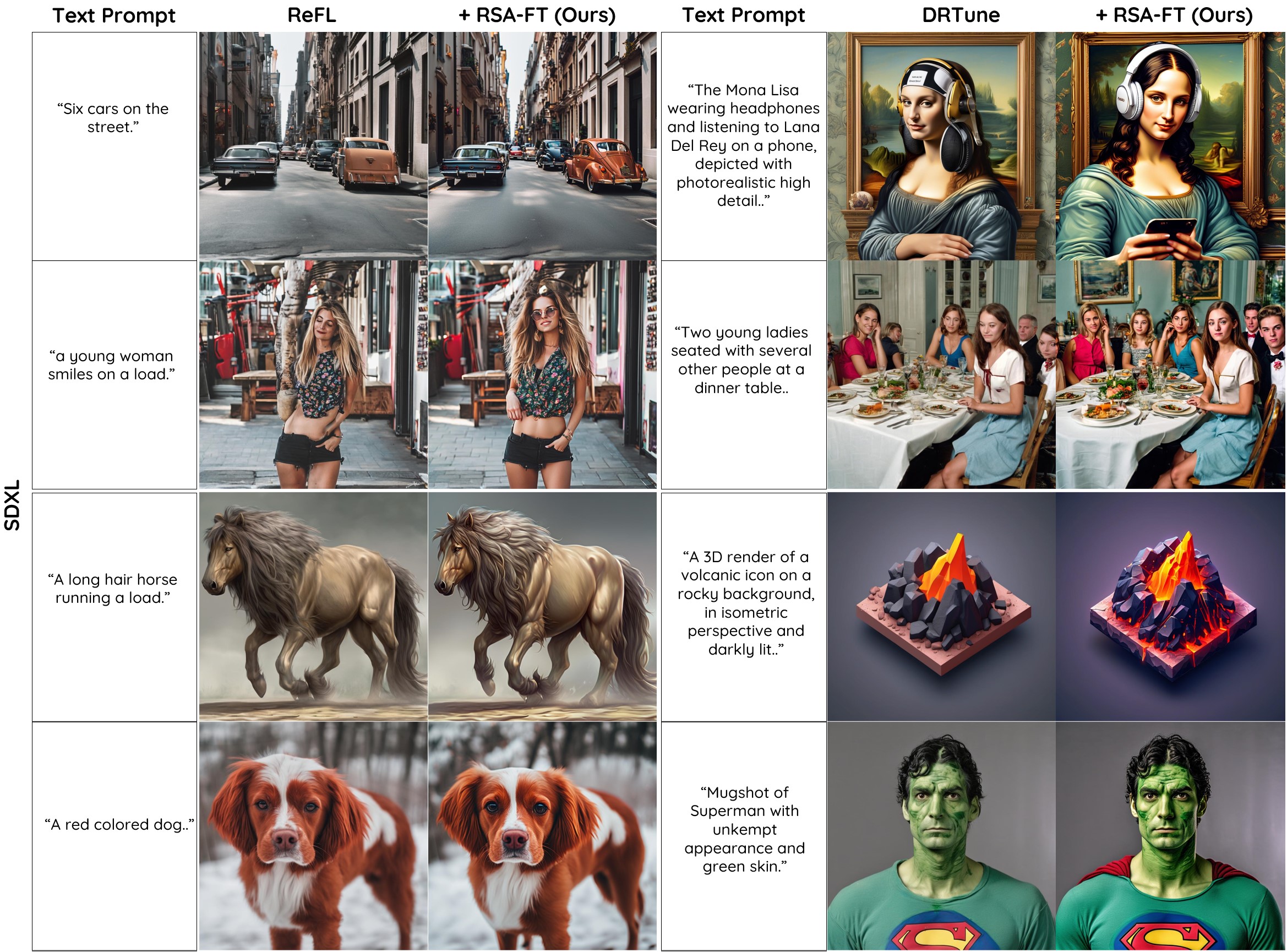}
\caption{
\textbf{Qualitative comparison on the SDXL backbone with and without our method.}
Each image is generated using the same text prompt and random seed across all methods.
Models equipped with our method (RSA-FT) produce images with more accurate text–image alignment and higher visual quality compared to their baselines. Moreover, RSA-FT suppresses the unwanted structural artifacts seen in the baselines.
}
\label{fig:fig_comparison2}
\vspace{-1em}
\end{figure*}

\begin{figure*}[t!]
\centering
\includegraphics[width=1\linewidth]{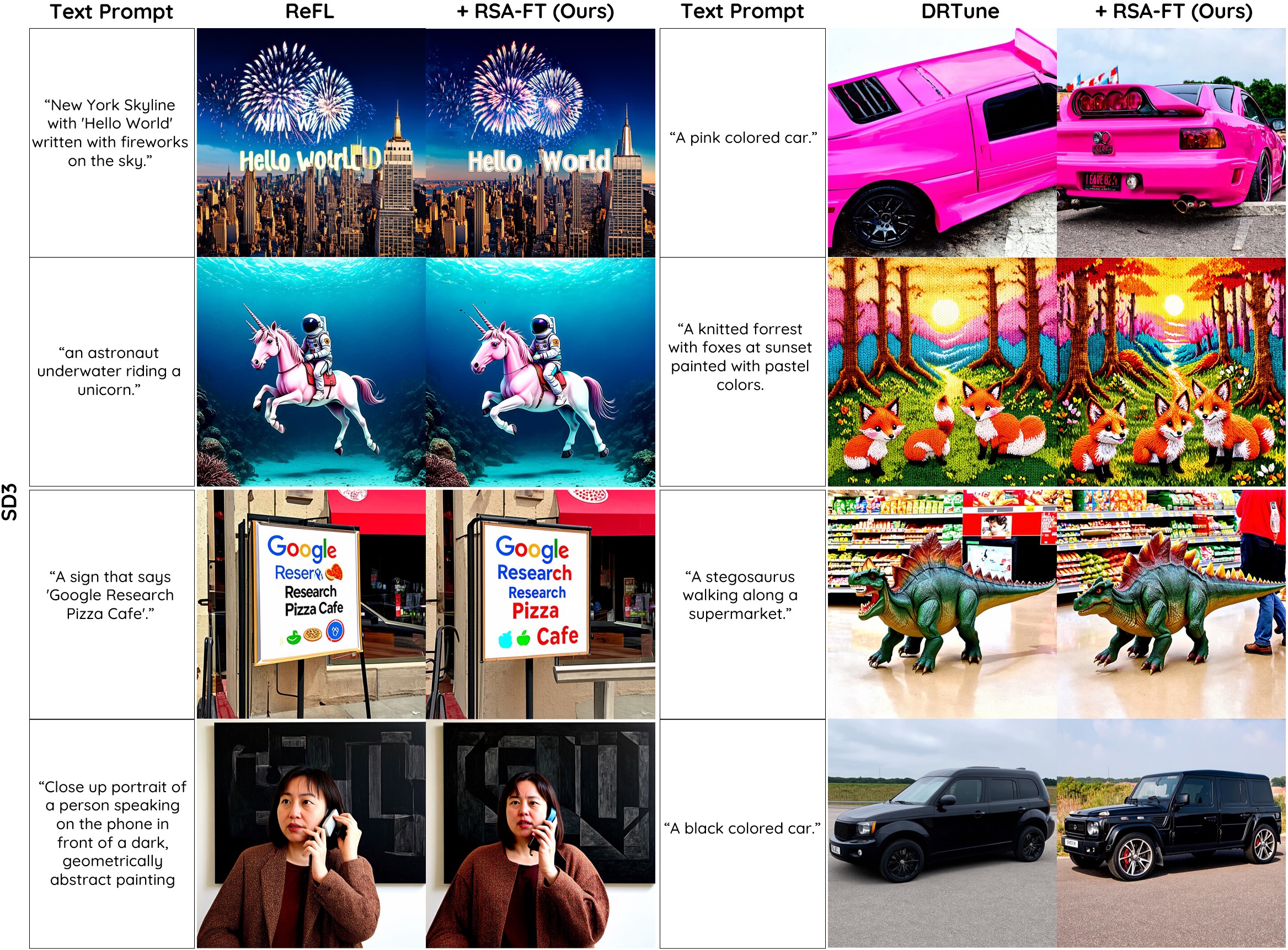}
\caption{
\textbf{Qualitative comparison on the SD3 backbone with and without our method.}
Each image is generated using the same text prompt and random seed across all methods.
Models equipped with our method (RSA-FT) produce images with more accurate text–image alignment and higher visual quality compared to their baselines. Moreover, RSA-FT preserves text rendering quality, introducing fewer distortions than the baseline methods.
}
\label{fig:fig_comparison3}
\vspace{-1em}
\end{figure*}

Figures~\ref{fig:fig_comparison1}, \ref{fig:fig_comparison2}, and \ref{fig:fig_comparison3} provide additional qualitative examples across a wide range of backbones, including SD1.5, SDXL, and SD3. For each RDRL baseline, we compare results with and without our proposed RSA-FT approach.

Across diverse prompts and model families, RSA-FT consistently improves the overall visual quality as perceived by humans, producing more coherent and stable images. In addition, many reward-induced artifacts commonly observed in RDRL pipelines are substantially reduced when applying RSA-FT. These qualitative examples demonstrate that our method strengthens both the generality and robustness of reward-driven diffusion fine-tuning, complementing the quantitative gains reported in the main paper.

\clearpage

\end{document}